\documentclass[times,final,3p]{elsarticle}

\usepackage{amssymb}
\usepackage{amsmath}
\usepackage{bm}
\usepackage{array}
\usepackage{booktabs}
\usepackage{multirow}
\usepackage{url}
\usepackage{float}
\biboptions{sort&compress}

\begin{document}

\begin{frontmatter}

\title{Structure-preserving uncertainty quantification for GENERIC dynamics}

\author[label1]{Zequn He\corref{cor1}}\ead{hezequn@engineering.upenn.edu}
\author[label1]{Celia Reina\corref{cor1}} \ead{creina@engineering.upenn.edu}
\affiliation[label1]{organization={Department of Mechanical Engineering and Applied Mechanics, University of Pennsylvania},
            city={Philadelphia},
            postcode={19104},
            state={PA},
            country={United States of America}}

\cortext[cor1]{Corresponding authors.}

\begin{abstract}
Structure-preserving machine learning embeds physical structure directly into model architectures, yet uncertainty quantification (UQ) for such hard-constrained models remains limited because standard UQ methods may violate the encoded admissibility conditions, require architectural modifications, or impose substantial computational costs. In this work, we propose Structure-Preserving Epistemic Neural Networks (S-PENNs), a general framework for UQ in scientific machine learning models with hard architectural constraints, and instantiate it for GENERIC (General Equation for Non-Equilibrium Reversible-Irreversible Coupling) dynamics. S-PENNs preserve the structural constraints of a pretrained model by attaching lightweight epinets to its constrained components, ensuring that every sampled realization remains physically admissible by construction. When applied to GENERIC dynamics, such a proposed framework yields thermodynamically consistent rollouts that preserve the first and second laws. Furthermore, we combine S-PENNs with split conformal prediction as a post-hoc calibration method to produce prediction intervals with finite-sample marginal coverage guarantees. We validate S-PENNs on three numerical examples: a harmonic oscillator coupled to a heat bath and an idealized chemical motor, both governed by ODEs, and a one-dimensional viscoplastic model governed by PDEs. Across all three examples, S-PENNs produce thermodynamically consistent 
stochastic realizations and well-calibrated prediction intervals while reducing the computational cost by about $1-3$ orders of magnitude compared to deep ensembles. Although the present study focuses on GENERIC dynamics, S-PENNs can be extended more broadly to scientific machine learning models in computational mechanics with either hard or soft constraints.
\end{abstract}


\end{frontmatter}


\section{Introduction}
\label{sec:intro}
Scientific machine learning has become an important computational tool in applied mechanics, where learned models are used to approximate governing equations, constitutive laws, and solution operators from data and physical knowledge. Representative subclasses include neural ODEs for continuous-time dynamics~\cite{chen2018neural}, physics-informed neural networks (PINNs) for differential-equation-constrained learning~\cite{raissi2019physics, karniadakis2021physics}, and neural operators for maps between function spaces~\cite{lu2021learning, li2020fourier, kovachki2023neural}. A central distinction among these methods is how physical constraints are imposed. In soft-constrained formulations, the governing equations or admissibility conditions enter the training objective as residual or penalty terms. In hard-constrained formulations, often referred to as structure-preserving machine learning, the model class is restricted through the architecture or parameterization itself, so selected invariants, stability properties, or thermodynamic laws are satisfied by construction rather than only encouraged during optimization~\cite{celledoni2021structure, jin2020sympnets, yu2021onsagernet, zhang2022gfinns}.

Within this structure-preserving paradigm, reversible dynamics have served as the initial proving ground. Hamiltonian Neural Networks~\cite{greydanus2019hamiltonian} and Lagrangian Neural Networks~\cite{cranmer2019lagrangian} infer the dynamics from learned energy or action principles, whereas SympNets~\cite{jin2020sympnets}, Symplectic ODE-Net~\cite{zhong2020symplectic}, and Poisson neural networks~\cite{jin2022learning} encode symplectic or Poisson geometry directly in the model class. However, many mechanical systems are governed by coupled reversible and irreversible dynamics. The GENERIC (General Equation for Non-Equilibrium Reversible-Irreversible Coupling) formalism~\cite{grmela1997dynamics, ottinger1997dynamics, ottinger2005beyond, pavelka2018multiscale}, which is tightly connected to metriplectic evolution~\cite{morrison1986paradigm}, provides a natural foundation for such systems by decomposing the evolution into reversible and irreversible contributions, subject to 
structural conditions that imply energy conservation and nonnegative entropy production. These thermodynamic guarantees make GENERIC a natural inductive bias for machine learning where neural networks provide flexible parameterizations of the thermodynamic building blocks, while the GENERIC structure controls their physical admissibility. Early work in this direction adopted soft constraints, as in Structure-Preserving Neural Networks, which learn conservative-dissipative dynamics by adding the so-called degeneracy conditions to the training objective~\cite{hernandez2021structure}. In contrast, hard-constrained formulations build the structural conditions into the parameterization itself and have focused largely on the linear GENERIC/metriplectic setting, where the irreversible contribution is represented by a friction or dissipative operator applied to the entropy gradient. In this setting, bracket-based models parameterize admissible dissipative brackets~\cite{lee2021machine}. GFINNs reparameterize the reversible and irreversible operators so the GENERIC degeneracy conditions are satisfied for deterministic and stochastic systems~\cite{zhang2022gfinns}. Neural Metriplectic Systems provide superior scalability of metriplectic dynamics that retains the energy-entropy structure~\cite{gruber2025efficiently}. Nonlinear GENERIC-Embedded Neural Networks (N-GENNs) move beyond this linear setting by learning nonlinear GENERIC dynamics with general dissipation potentials, including non-quadratic ones~\cite{votruba2026nonlinear}, combining GFINN-style projections for the reversible degeneracy condition with a convex dissipation-potential parameterization as introduced in VONNs~\cite{huang2022variational}. 
These models secure thermodynamic admissibility by architectural design, but admissibility is not the same as predictive reliability. The predictions may satisfy the thermodynamic laws while still being poorly supported by sparse data, noisy measurements, model misspecification, or extrapolation beyond the training regime. To use these learned models in practice, one crucial requirement is to obtain uncertainty estimates that preserve the imposed structure while quantifying predictive reliability.

Uncertainty quantification (UQ) provides the standard framework for characterizing such predictive uncertainty and has received substantial attention in scientific machine learning. In broad terms, uncertainty in learned surrogates is commonly separated into aleatoric uncertainty, which captures irreducible variability from noise or unresolved stochasticity, and epistemic uncertainty, which captures lack of knowledge due to limited data, model misspecification, or unobserved regimes~\cite{psaros2023uncertainty}. Standard UQ methods include Bayesian neural networks (BNNs)~\cite{neal2012bayesian,goan2020bayesian}, which place priors over network parameters and infer a posterior predictive distribution. Deep ensembles~\cite{lakshminarayanan2017simple} estimate uncertainty from the spread of predictions across independently initialized and trained models. Monte Carlo dropout~\cite{gal2016dropout} keeps dropout active at test time so that repeated stochastic forward passes are interpreted as samples from an approximate variational posterior. In physics-informed machine learning, existing UQ methods can be organized by where uncertainty is introduced. Bayesian PINNs (B-PINNs) place priors on network parameters (or unknown physical parameters in inverse problems) and infer posterior predictive distributions, for example, with Hamiltonian Monte Carlo or variational inference~\cite{yang2021b}. 
NN-aPC uses deep NNs to learn modal functions of arbitrary Polynomial Chaos (aPC) expansion of the solution or parameters in stochastic PDEs, and employs dropout to quantify their uncertainty~\cite{zhang2019quantifying}.
Adversarial UQ PINNs use latent-variable generative models to represent PDE solution distributions under random physical inputs (e.g., boundary or initial data) or noisy measurements, while WGAN-PINNs learn uncertainty in initial or boundary data and propagate it to interior solution fields through physics-informed constraints~\cite{yang2019adversarial,gao2022wasserstein}. 
Physics-informed variational autoencoders (PI-VAE) and physics-informed variational inference introduce latent variables and physics-based constraints to solve forward and inverse problems involving stochastic differential equations~\cite{zhong2023pi,shin2023physics}.
Physics-informed polynomial chaos expansions augment conventional PCE surrogate construction with differential-equation and boundary-condition constraints~\cite{novak2024physics}. More recently, conformal prediction (CP) has been used to calibrate existing PINN uncertainty estimators into intervals with finite-sample coverage guarantees and spatial adaptivity~\cite{yu2026conformal}. These approaches target soft-penalized or physics-regularized surrogates, rather than models with hard-encoded physical constraints in which every sampled realization must satisfy structural admissibility conditions.

By contrast, UQ for models with hard-encoded physical constraints must preserve admissibility rather than merely estimate variation around a physics-regularized predictor. Existing methods have demonstrated such a requirement in several specialized settings. For Hamiltonian systems with additive dissipation, Symplectic Spectrum Gaussian Processes (SSGP) use a symplectic Gaussian-process prior with random Fourier features to infer Hamiltonians from noisy, sparse data while retaining conservative or dissipative structure~\cite{tanaka2022symplectic}. Similarly, Hamiltonian Gaussian Processes (HGP) use a decoupled Gaussian-process Hamiltonian and energy-conserving shooting for inference from long noisy trajectories~\cite{ross2023learning}, while Bayesian identification of nonseparable Hamiltonians combines a structure-preserving Hamiltonian parameterization, dependent additive and multiplicative noise models, and reduced-order Bayesian inference~\cite{galioto2024bayesian}. Thermodynamic and constitutive examples use analogous admissible representations. Thermodynamically constrained Gaussian-process equations of state capture model and data uncertainty 
subject to consistency and stability constraints~\cite{sharma2024learning}. Bayesian-EUCLID performs Bayesian sparse discovery over a hyperelastic feature library from displacement and reaction-force data, quantifying uncertainty in active constitutive terms and data noise through a momentum-balance likelihood and spike-slab prior~\cite{joshi2022bayesian}. Bayesian constitutive artificial neural networks learn probability distributions for the weights of free-energy-based constitutive networks, yielding credible intervals for the used model terms and associated 
constitutive responses~\cite{linka2025discovering}. More recently, conformal quantile regression adopts a frequentist approach to probabilistic modeling and has calibrated tensor-valued quantile predictions from a strain-invariant, polyconvex, and thermodynamically consistent constitutive network, with the resulting intervals targeting aleatoric uncertainty from data variability rather than epistemic uncertainty~\cite{bahmani2026conformal}. This body of work shows that UQ for hard-encoded physics constraints is both desirable and feasible, but it also indicates that generic UQ methods do not automatically yield hard-constrained UQ easily.

The core limitation is that hard-encoded constraints restrict the admissible model class itself, so sampled UQ realizations must remain in a space that is closed under those constraints. Randomizing unconstrained weights or perturbing intermediate activations can move a realization outside that space, while preserving admissibility by independent replication requires training complete constrained models. BNNs can preserve structure only when the prior, posterior approximation, and inference procedure are defined over the constrained surrogate itself; otherwise, generic weight-space uncertainty does not enforce the required relations, and posterior inference over a constrained architecture can be expensive at scale~\cite{jospin2022hands}. Deep ensembles can preserve constraints by training several full constrained models with cost growing essentially linearly in the ensemble size~\cite{lakshminarayanan2017simple, gawlikowski2023survey}. MC dropout uses stochastic internal masks, so hard-encoded properties may be lost unless every masked subnetwork still satisfies the same constraints~\cite{gal2016dropout, gawlikowski2023survey}. This difficulty increases in structure-preserving models whose admissibility usually depends on several interacting constrained building blocks rather than on a single output map. Therefore, a practical UQ framework is expected to introduce uncertainty inside the constrained parameterization, preserve the structural relations under each realization, propagate uncertainty coherently through the coupled constrained blocks, and keep the added computational cost tractable.

To address the gap and meet the requirements, we propose Structure-Preserving Epistemic Neural Networks (S-PENNs), an epistemic UQ framework that injects uncertainty into constrained scientific machine learning parameterizations without violating their encoded physical and structural relations. S-PENNs build on the Epistemic Neural Networks (ENNs) introduced in~\cite{osband2023epistemic}, in which a conventional base network is augmented by a lightweight auxiliary network called the epinet. Recent work has applied this idea to operator learning through NEON~\cite{guilhoto2024composite}, to PINNs through E-PINNs~\cite{jacob2025pinns}, and to thermodynamics-informed diffusion models through EVODMs~\cite{he2026evodms} and SPIEDiff~\cite{he2025spiediff}. Nevertheless, none of these works address structure-preserving architectures formally in settings in which each UQ realization must satisfy coupled admissibility conditions by construction. To demonstrate the feasibility of S-PENNs, we develop and instantiate the framework for GENERIC dynamics, specifically, on top of the N-GENNs~\cite{votruba2026nonlinear}. This is a particularly demanding test bed for structure-preserving UQ because the GENERIC formalism couples four thermodynamic building blocks through skew-symmetry, convexity, and degeneracy conditions that jointly enforce the first and second laws of thermodynamics, which will be detailed in Section~\ref{sec:N-GENNs_intro}. Any UQ method equipped on N-GENNs must preserve these constraints jointly rather than one block at a time, and the perturbations must propagate coherently across blocks to yield consistent uncertainty estimates for the resulting dynamics. The key idea here for S-PENNs is to treat the deterministic N-GENNs as base networks and attach a separate epinet to each building block of N-GENNs so that each epinet inherits the architectural properties of its corresponding block. Since the perturbed blocks are then assembled through the same structure-preserving reparameterizations as in N-GENNs, every sampled realization remains thermodynamically admissible by construction. The component-wise perturbations are driven by a common source of randomness. Each epinet is conditioned on stop-gradient features from its corresponding deterministic block and on the same inputs as that block, while the shared epistemic index couples the block-wise perturbations across the assembled GENERIC vector field. One uncertainty realization therefore perturbs the entire GENERIC parameterization coherently and propagates uncertainty consistently across blocks without breaking the thermodynamic structure. The nonlinear GENERIC instantiation of S-PENNs also requires uncertainty representations for global, input-independent quantities, such as the state-independent terms in the GENERIC reparameterization. This setting is analogous to inverse problems with unknown physical parameters, where the parameters are inferred jointly with the solution field but are not themselves functions of the input variables. Prior epinet-based work in scientific machine learning has either not considered such global parameters~\cite{guilhoto2024composite, he2026evodms, he2025spiediff} or handled them indirectly by softly penalizing parameter variance during optimization~\cite{jacob2025pinns}. To have a hard-constrained solution for this problem, we introduce an epinet construction tailored to global parameters, which enforces input-independent uncertainty by construction with no soft-penalized variance term. Lastly, beyond preserving thermodynamic consistency, we use split conformal prediction~\cite{angelopoulos2023conformal} on S-PENNs as a post-hoc calibration step to obtain distribution-free finite-sample marginal coverage guarantees on the resulting prediction intervals. The proposed framework is evaluated on three numerical examples: a harmonic oscillator coupled to a heat bath, an idealized chemical motor, and a one-dimensional viscoplastic model. Across all three examples, S-PENNs yield thermodynamically admissible uncertainty realizations and well-calibrated prediction intervals while reducing training cost by about $1-3$ orders of magnitude compared to deep ensembles. Although the presented work focuses on GENERIC dynamics, the proposed idea extends to broader classes of scientific machine learning models that encode physical structure through either hard architectural constraints or soft physics-informed penalties.

The remainder of the paper is organized as follows. Section~\ref{sec:method} reviews the GENERIC formalism and the N-GENNs framework of~\cite{votruba2026nonlinear}, introduces the S-PENNs construction, and presents the conformal calibration procedure. Section~\ref{sec:numerical_experiments} reports the main numerical results for a harmonic oscillator coupled to a heat bath (Section~\ref{sec:harmonic_oscillator}), an idealized chemical motor (Section~\ref{sec:chemical_motor}), and a one-dimensional viscoplastic model (Section~\ref{sec:viscoplastic_model}). Section~\ref{sec:conclusion} discusses limitations and future directions. Additional inverse-problem benchmarks demonstrating the proposed epinet construction are provided in~\ref{sec:epinet_inverse_problems}.

\section{Methodology}
\label{sec:method}
This section begins by reviewing the structure-preserving N-GENNs framework for nonlinear GENERIC dynamics in Section~\ref{sec:N-GENNs_intro}. Section~\ref{sec:enns_intro} reviews the epinet construction, and Section~\ref{sec:s-penns} specializes it to N-GENNs to obtain the proposed S-PENNs architecture. Finally, Section~\ref{sec:cp_calibration} presents the split conformal calibration procedure used to obtain finite-sample marginal coverage guarantees for the resulting prediction intervals.

\subsection{Nonlinear GENERIC-embedded neural networks (N-GENNs)}
\label{sec:N-GENNs_intro}
We first recall the nonlinear GENERIC formalism, which decomposes the evolution into a reversible Hamiltonian part and an irreversible generalized gradient flow. For a finite-dimensional state vector $\mathbf{x}\in\mathbb{R}^d$ and its conjugate variable $\mathbf{x}^{*}$, the nonlinear GENERIC dynamics are written as
\begin{equation}
    \label{eq:nonlinear_GENERIC_eqn}
    \dot{\mathbf{x}} = L(\mathbf{x})\mathrm{D} E(\mathbf{x}) + \left.\mathrm{D}_{\mathbf{x}^{*}}\Xi(\mathbf{x},\mathbf{x}^{*})\right|_{\mathbf{x}^{*}=\mathrm{D} S(\mathbf{x})}.
\end{equation}
Here $L$ is the Poisson operator describing reversible dynamics, $E$ is the total energy, $S$ is the entropy, and $\Xi$ is the dissipation potential. In finite-dimensional settings, $\mathrm{D}$ and $\mathrm{D}_{\mathbf{x}^{*}}$ denote ordinary partial derivatives; in infinite-dimensional settings, where the state variables are fields, they denote the corresponding functional derivatives. Among the thermodynamic constraints imposed on the GENERIC building blocks, the two degeneracy conditions are
\begin{equation}
    \label{eq:N-GENNs_constraints}
    \begin{aligned}
    &\text{reversible degeneracy condition} &&
    L(\mathbf{x})\mathrm{D} S(\mathbf{x})=\mathbf{0}, \quad \forall \mathbf{x},\\
    &\text{irreversible degeneracy condition} &&
    \Xi(\mathbf{x},\mathbf{x}^{*}+\lambda \mathrm{D}E(\mathbf{x}))=\Xi(\mathbf{x},\mathbf{x}^{*}),
    \quad \forall \mathbf{x},\mathbf{x}^{*}, \quad \forall \lambda\in\mathbb{R}.
    \end{aligned}
\end{equation}
When combined with the skew-symmetry of the Poisson operator, $L(\mathbf{x})=-L(\mathbf{x})^{\top}$, and the admissibility conditions $\Xi(\mathbf{x},\mathbf{0})=0$, $\Xi(\mathbf{x},\mathbf{x}^{*})\ge 0$, and convexity of $\Xi$ with respect to $\mathbf{x}^{*}$, these degeneracy conditions ensure conservation of total energy and nonnegative entropy production~\cite{grmela2018generic, kraaij2020fluctuation}.

Building on the above GENERIC structure, N-GENNs~\cite{votruba2026nonlinear} represent the thermodynamic building blocks with neural networks and enforce the constraints through structure-preserving reparameterizations. With the deterministic trainable parameter set denoted by $\bm{\psi}$\footnote{In this subsection, $\bm{\psi}$ denotes the aggregate trainable parameter set of the deterministic N-GENNs backbone. It includes neural-network weights and biases as well as finite-dimensional trainable variables used in the structure-preserving reparameterizations. A shared subscript $\bm{\psi}$ is used here only for clarity and does not imply weight sharing among distinct networks.}, the deterministic backbone uses scalar networks $E_{\bm{\psi}}$ and $S_{\bm{\psi}}$, a reversible operator network $\tilde{L}_{\bm{\psi}}$, and a raw dissipation-potential network $\tilde{\Xi}_{\bm{\psi}}$. Let $\bm{B}_{\bm{\psi}}=(\bm{B}_{1\bm{\psi}},\dots,\bm{B}_{d\bm{\psi}})\in\mathbb{R}^{d\times d\times d}$ denote the trainable matrices included in $\bm{\psi}$, and define
\begin{equation}
    \bm{A}_{i\bm{\psi}}=\bm{B}_{i\bm{\psi}} - \bm{B}_{i\bm{\psi}}^{\top}, \qquad i=1,\dots,d.
\end{equation}
The matrix $Q_{S_{\bm{\psi}}}(\mathbf{x})\in\mathbb{R}^{d\times d}$ is constructed so that its $i$th row is $(\bm{A}_{i\bm{\psi}}\mathrm{D} S_{\bm{\psi}}(\mathbf{x}))^{\top}$, and the reparameterized Poisson operator is defined as
\begin{equation}
    \label{eq:n-genn_poisson_operator_reparam}
    L_{\bm{\psi}}(\mathbf{x})=Q_{S_{\bm{\psi}}}(\mathbf{x})^{\top}\tilde{L}_{\bm{\psi}}(\mathbf{x})Q_{S_{\bm{\psi}}}(\mathbf{x}).
\end{equation}
Because each $\bm{A}_{i\bm{\psi}}$ is skew-symmetric, $Q_{S_{\bm{\psi}}}(\mathbf{x})\mathrm{D}S_{\bm{\psi}}(\mathbf{x})=0$, which gives $L_{\bm{\psi}}(\mathbf{x})\mathrm{D}S_{\bm{\psi}}(\mathbf{x})=0$. This enforces the reversible degeneracy condition by construction.

For the irreversible part, 
the raw dissipation potential $\tilde{\Xi}_{\bm{\psi}}(\mathbf{x},\mathbf{x}^{*})$ is represented by a partially input-convex neural network (PICNN)~\cite{amos2017input, huang2022variational}, treating $\mathbf{x}^{*}$ as the convex variable and $\mathbf{x}$ as a conditioning input. Let $P_{E_{\bm{\psi}}}$ be the projection matrix onto the orthogonal complement of $\mathrm{D}E_{\bm{\psi}}$,
\begin{equation}
    P_{E_{\bm{\psi}}}(\mathbf{x})=I - \frac{\mathrm{D}E_{\bm{\psi}}(\mathbf{x})\mathrm{D}E_{\bm{\psi}}(\mathbf{x})^{\top}}
                            {\|\mathrm{D}E_{\bm{\psi}}(\mathbf{x})\|_2^2}.
\end{equation}
The dissipation potential used in the dynamics is obtained by the reparameterization as follows
\begin{equation}
    \label{eq:n-ginn_xi_reparam}
    \begin{aligned}
    \Xi_{\bm{\psi}}(\mathbf{x},\mathbf{x}^{*})
    &=
    \tilde{\Xi}_{\bm{\psi}}\!\left(\mathbf{x},P_{E_{\bm{\psi}}}(\mathbf{x})\mathbf{x}^{*}\right)
    -
    \tilde{\Xi}_{\bm{\psi}}(\mathbf{x},\mathbf{0})
    -
    \left(
    \left.
        \mathrm{D}_{\mathbf{x}^{*}}\tilde{\Xi}_{\bm{\psi}}(\mathbf{x},\mathbf{x}^{*})
    \right|_{\mathbf{x}^{*}=\mathbf{0}}
    \right)^{\top}
    P_{E_{\bm{\psi}}}(\mathbf{x})\mathbf{x}^{*}.
    \end{aligned}
\end{equation}
Here, $\mathbf{x}^{*}$ enters only through $P_{E_{\bm{\psi}}}(\mathbf{x})\mathbf{x}^{*}$. Since $P_{E_{\bm{\psi}}}(\mathbf{x})\mathrm{D}E_{\bm{\psi}}(\mathbf{x})=0$, replacing $\mathbf{x}^{*}$ by $\mathbf{x}^{*}+\lambda\mathrm{D}E_{\bm{\psi}}(\mathbf{x})$ leaves this projected argument unchanged, and therefore $\Xi_{\bm{\psi}}(\mathbf{x},\mathbf{x}^{*}+\lambda\mathrm{D}E_{\bm{\psi}}(\mathbf{x}))=\Xi_{\bm{\psi}}(\mathbf{x},\mathbf{x}^{*})$. As a result, the irreversible degeneracy condition is satisfied. The reparameterization preserves convexity in $\mathbf{x}^{*}$ because it combines a linear composition of the raw convex potential with an affine subtraction. The affine correction further gives $\Xi_{\bm{\psi}}(\mathbf{x},\mathbf{0})=0$ and $\mathrm{D}_{\mathbf{x}^{*}}\Xi_{\bm{\psi}}(\mathbf{x},\mathbf{0})=0$, so by convexity $\mathbf{x}^{*}=\mathbf{0}$ is a global minimizer and $\Xi_{\bm{\psi}}(\mathbf{x},\mathbf{x}^{*})\ge 0$. These properties make $\Xi_{\bm{\psi}}$ admissible by construction.

The constrained building blocks define the deterministic N-GENNs vector field
\begin{equation}
    \label{eq:deterministic_GENERICNNs}
    g_{\bm{\psi}}(\mathbf{x})=L_{\bm{\psi}}(\mathbf{x})\mathrm{D} E_{\bm{\psi}}(\mathbf{x})+\left.\mathrm{D}_{\mathbf{x}^{*}}\Xi_{\bm{\psi}}(\mathbf{x},\mathbf{x}^{*})\right|_{\mathbf{x}^{*}=\mathrm{D} S_{\bm{\psi}}(\mathbf{x})}.
\end{equation}
Since the reversible and irreversible constraints are built into the parameterization, $g_{\bm{\psi}}$ is thermodynamically admissible by construction.

\subsection{Epistemic neural networks and the epinet}
\label{sec:enns_intro}
According to~\cite{osband2023epistemic}, ENNs refer to conventional neural networks augmented with a lightweight auxiliary network, the epinet. Such models can be written as
\begin{equation}
    \label{eq:enn_general}
    \tau_{\bm{\vartheta}}(\mathbf{x},\mathbf{z})=\mu_{\bm{\psi}}(\mathbf{x})+
    \sigma_{\bm{\phi}}(\bar{\bm{h}}_{\bm{\psi}}(\mathbf{x}),\mathbf{z}),
    \qquad
    \bar{\bm{h}}_{\bm{\psi}}(\mathbf{x})=
    [\mathrm{sg}\!\left(\bm{h}_{\bm{\psi}}(\mathbf{x})\right),\mathbf{x}],
\end{equation}
where $\bm{\psi}$ and $\bm{\phi}$ denote the base-network and epinet trainable parameter sets, respectively, and $\bm{\vartheta}=(\bm{\psi},\bm{\phi})$ denotes the trainable parameter set of the augmented model. Here $\mu_{\bm{\psi}}(\mathbf{x})$ denotes the base network and $\sigma_{\bm{\phi}}(\bar{\bm{h}}_{\bm{\psi}}(\mathbf{x}),\mathbf{z})$ is the epinet. $\mathbf{z}\in\mathbb{R}^{d_z}$ is an epistemic index drawn from a chosen reference distribution $\pi(\mathbf{z})$. $\bm{h}_{\bm{\psi}}(\mathbf{x})$ represents the base-network features, typically taken from the last hidden layer, and $\mathrm{sg}(\cdot)$ is the stop-gradient operator that prevents gradients from flowing into the base network. Throughout the paper, the notation $\bar{\bm{h}}_{\bm{\psi}}(\mathbf{x})=[\mathrm{sg}(\bm{h}_{\bm{\psi}}(\mathbf{x})),\mathbf{x}]$ means the concatenation of stop-gradient hidden features and the raw input variables. This is the same typical setting used in~\cite{osband2023epistemic}, but we state it explicitly here to avoid ambiguity.

The epinet is decomposed into a learnable network $\sigma_{\bm{\phi}}^{\mathrm{learn}}$ and a prior network $\sigma^{\mathrm{prior}}$,
\begin{equation}
    \label{eq:epinet_decomposition}
    \sigma_{\bm{\phi}}(\bar{\bm{h}},\mathbf{z})=
    \sigma_{\bm{\phi}}^{\mathrm{learn}}(\bar{\bm{h}},\mathbf{z})
    +
    w\,\sigma^{\mathrm{prior}}(\bar{\bm{h}},\mathbf{z}),
\end{equation}
where $w>0$ is the prior-weight hyperparameter that scales the contribution of the fixed random prior network relative to the learnable network, thereby controlling the strength of the prior perturbation and hence the initial epistemic spread. Since the prior network is randomly initialized and kept fixed throughout training, its parameters are omitted from the notation for conciseness. The learnable and prior networks are written as
\begin{align}
    \sigma_{\bm{\phi}}^{\mathrm{learn}}(\bar{\bm{h}},\mathbf{z})
    &=
    \mathrm{NN}_{\bm{\phi}}([\bar{\bm{h}},\mathbf{z}])^{\top}\mathbf{z},
    \label{eq:epinet_learnable_network}\\
    \sigma^{\mathrm{prior}}(\bar{\bm{h}},\mathbf{z})
    &=
    \sum_{n=1}^{d_{z}}z_n\,\sigma_n^{\mathrm{prior}}(\bar{\bm{h}}),
    \label{eq:epinet_prior_network}
\end{align}
respectively. Here $\mathrm{NN}_{\bm{\phi}}$ is a trainable network whose output dimension matches that of $\mathbf{z}$, and $\{\sigma_n^{\mathrm{prior}}\}_{n=1}^{d_{z}}$ are fixed randomly initialized networks that typically share the same architecture as the learnable network, with one prior network for each component of the epistemic index.

\subsection{Structure-preserving epistemic neural networks (S-PENNs)}
\label{sec:s-penns}
\begin{figure}[htbp]
    \centering
    \includegraphics[width=\textwidth]{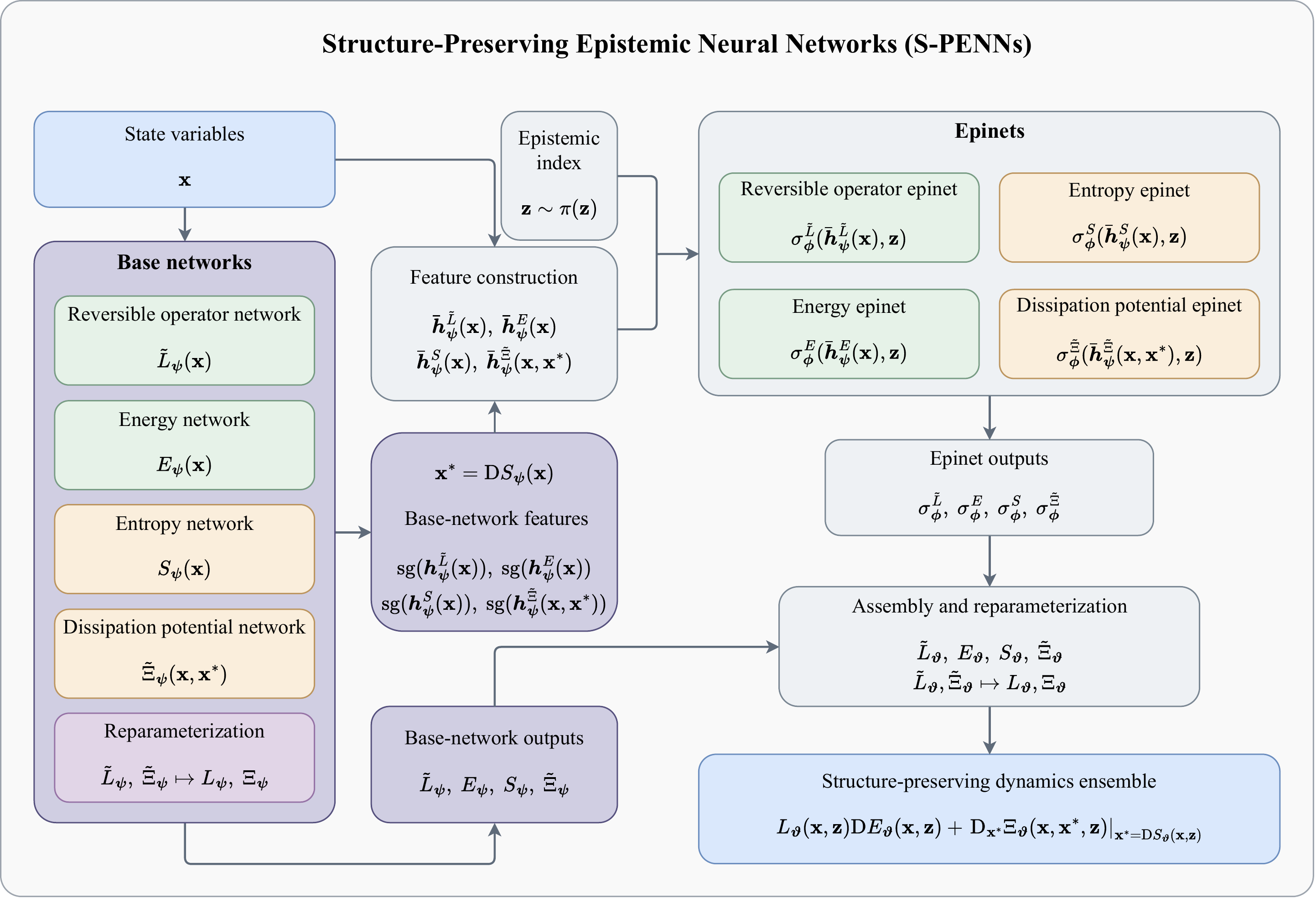}
    \caption{Schematic of the S-PENNs construction for GENERIC dynamics. The state variables $\mathbf{x}$ are passed through the deterministic N-GENNs base networks, which produce the raw thermodynamic building blocks $\tilde{L}_{\bm{\psi}}$, $E_{\bm{\psi}}$, $S_{\bm{\psi}}$, and $\tilde{\Xi}_{\bm{\psi}}$, together with their hidden representations. The stop-gradient hidden representations and the relevant inputs are used to construct epinet feature vectors $\bar{\bm{h}}_{\bm{\psi}}^{\tilde{L}}$, $\bar{\bm{h}}_{\bm{\psi}}^{E}$, $\bar{\bm{h}}_{\bm{\psi}}^{S}$, and $\bar{\bm{h}}_{\bm{\psi}}^{\tilde{\Xi}}$. These feature vectors, together with a shared epistemic index $\mathbf{z}\sim\pi(\mathbf{z})$, are passed to block-specific epinets, which generate perturbations of the reversible operator, energy, entropy, and dissipation-potential branches. The epinet outputs are then assembled with the corresponding base-network outputs to obtain $\tilde{L}_{\bm{\vartheta}}$, $E_{\bm{\vartheta}}$, $S_{\bm{\vartheta}}$, and $\tilde{\Xi}_{\bm{\vartheta}}$; the same structure-preserving reparameterizations used in N-GENNs map the raw reversible operator and dissipation-potential blocks to $L_{\bm{\vartheta}}$ and $\Xi_{\bm{\vartheta}}$. Hence each draw of $\mathbf{z}$ defines one admissible GENERIC vector field $L_{\bm{\vartheta}}(\mathbf{x},\mathbf{z})\mathrm{D}E_{\bm{\vartheta}}(\mathbf{x},\mathbf{z})+\left.\mathrm{D}_{\mathbf{x}^{*}}\Xi_{\bm{\vartheta}}(\mathbf{x},\mathbf{x}^{*},\mathbf{z})\right|_{\mathbf{x}^{*}=\mathrm{D}S_{\bm{\vartheta}}(\mathbf{x},\mathbf{z})}$, and repeated draws form a structure-preserving dynamics ensemble.}
    \label{fig:overview_framework}
\end{figure}

We now introduce S-PENNs and specialize the framework to N-GENNs for learning GENERIC dynamics with quantified uncertainty. S-PENNs attach epinets to the thermodynamic building blocks of the N-GENNs backbone, and reassemble the perturbed objects through the same structure-preserving parameterization. Consequently, every draw of the epistemic index yields a thermodynamically consistent dynamics sample. Fig.~\ref{fig:overview_framework} depicts the overall framework.

To preserve the thermodynamic structure under epistemic perturbations, we attach a separate epinet to each deterministic building block in Eq.~\eqref{eq:deterministic_GENERICNNs}, i.e., the reversible operator network $\tilde{L}_{\bm{\psi}}$, the energy network $E_{\bm{\psi}}$, the entropy network $S_{\bm{\psi}}$, and the raw dissipation-potential network $\tilde{\Xi}_{\bm{\psi}}$. The augmented blocks can be written as
\begin{equation}
    \label{eq:s-penn_building_blocks}
    \begin{aligned}
        \tilde{L}_{\bm{\vartheta}}(\mathbf{x},\mathbf{z})
        &=
        \tilde{L}_{\bm{\psi}}(\mathbf{x})
        +
        \sigma_{\bm{\phi}}^{\tilde{L}}(\bar{\bm{h}}_{\bm{\psi}}^{\tilde{L}}(\mathbf{x}),\mathbf{z}),\\
        E_{\bm{\vartheta}}(\mathbf{x},\mathbf{z})
        &=
        E_{\bm{\psi}}(\mathbf{x})
        +
        \sigma_{\bm{\phi}}^{E}(\bar{\bm{h}}_{\bm{\psi}}^{E}(\mathbf{x}),\mathbf{z}),\\
        S_{\bm{\vartheta}}(\mathbf{x},\mathbf{z})
        &=
        S_{\bm{\psi}}(\mathbf{x})
        +
        \sigma_{\bm{\phi}}^{S}(\bar{\bm{h}}_{\bm{\psi}}^{S}(\mathbf{x}),\mathbf{z}),\\
        \tilde{\Xi}_{\bm{\vartheta}}(\mathbf{x},\mathbf{x}^{*},\mathbf{z})
        &=
        \tilde{\Xi}_{\bm{\psi}}(\mathbf{x},\mathbf{x}^{*})
        +
        \sigma_{\bm{\phi}}^{\tilde{\Xi}}(\bar{\bm{h}}_{\bm{\psi}}^{\tilde{\Xi}}(\mathbf{x},\mathbf{x}^{*}),\mathbf{z}).
    \end{aligned}
\end{equation}
Here and below, the shared subscripts $\bm{\psi}$ and $\bm{\phi}$ denote aggregate parameter collections and do not imply weight sharing across thermodynamic blocks. The inputs to the correspoding epinet branches are denoted as $\bar{\bm{h}}_{\bm{\psi}}^{\tilde{L}}$, $\bar{\bm{h}}_{\bm{\psi}}^{E}$, $\bar{\bm{h}}_{\bm{\psi}}^{S}$, and $\bar{\bm{h}}_{\bm{\psi}}^{\tilde{\Xi}}$, more specifically, they are constructed as
\begin{equation}
    \label{eq:s-penn_feature_concat}
    \begin{aligned}
        \bar{\bm{h}}_{\bm{\psi}}^{\tilde{L}}(\mathbf{x})
        &=
        [\mathrm{sg}\!\left(\bm{h}_{\bm{\psi}}^{\tilde{L}}(\mathbf{x})\right),\mathbf{x}],\\
        \bar{\bm{h}}_{\bm{\psi}}^{E}(\mathbf{x})
        &=
        [\mathrm{sg}\!\left(\bm{h}_{\bm{\psi}}^{E}(\mathbf{x})\right),\mathbf{x}],\\
        \bar{\bm{h}}_{\bm{\psi}}^{S}(\mathbf{x})
        &=
        [\mathrm{sg}\!\left(\bm{h}_{\bm{\psi}}^{S}(\mathbf{x})\right),\mathbf{x}],\\
        \bar{\bm{h}}_{\bm{\psi}}^{\tilde{\Xi}}(\mathbf{x},\mathbf{x}^{*})
        &=
        [\mathrm{sg}\!\left(\bm{h}_{\bm{\psi}}^{\tilde{\Xi}}(\mathbf{x},\mathbf{x}^{*})\right),\mathbf{x},\mathbf{x}^{*}].
    \end{aligned}
\end{equation}
The same epistemic index $\mathbf{z}$ is used in all augmented blocks, so each draw selects a joint perturbation of the full GENERIC parameterization.

\subsubsection{Input-independent epinet branch for global tensors}
\label{sec:epinet_input_independent}
As introduced in Section~\ref{sec:N-GENNs_intro}, the N-GENNs backbone contains finite-dimensional trainable quantities that are not state-dependent neural-network outputs. The tensor $\bm{B}_{\bm{\psi}}=(\bm{B}_{1\bm{\psi}},\dots,\bm{B}_{d\bm{\psi}})\in\mathbb{R}^{d\times d\times d}$ is one such quantity. It is shared over the state space and enters the reversible operator reparameterization only through the skew matrices $\bm{A}_{i\bm{\psi}}$. For each fixed epistemic index, the perturbation should be a single state-independent tensor whereas a standard epinet does not enforce this property, and soft penalizing~\cite{jacob2025pinns} would only encourage, rather than guarantee, input independence. Alternatively, we propose to use a linear epinet branch driven only by the epistemic index,
\begin{equation}
    \label{eq:n-genn_global_matrix_head}
    \begin{aligned}
        \bm{B}_{\bm{\vartheta}}(\mathbf{z})
        &=
        \bm{B}_{\bm{\psi}}
        +
        \bm{\sigma}_{\bm{\phi}}^{B}(\mathbf{z}),\\
        \bm{\sigma}_{\bm{\phi}}^{B}(\mathbf{z})
        &=
        \bm{\sigma}_{\bm{\phi}}^{B,\mathrm{learn}}(\mathbf{z})
        +
        w\,\bm{\sigma}^{B,\mathrm{prior}}(\mathbf{z}),\\
        \bm{\sigma}_{\bm{\phi}}^{B,\mathrm{learn}}(\mathbf{z})
        &=
        \sum_{n=1}^{d_z} z_n\,\bm{\phi}_n^B,
        \qquad
        \bm{\sigma}^{B,\mathrm{prior}}(\mathbf{z})
        =
        \sum_{n=1}^{d_z} z_n\,\bm{\zeta}_n^B .
    \end{aligned}
\end{equation}
Here $\bm{\phi}_n^B,\bm{\zeta}_n^B\in\mathbb{R}^{d\times d\times d}$ have the same shape as $\bm{B}_{\bm{\psi}}$. The tensors $\bm{\phi}^B=\{\bm{\phi}_n^B\}_{n=1}^{d_z}$ are included in the epinet parameter set $\bm{\phi}$, whereas $\bm{\zeta}^B=\{\bm{\zeta}_n^B\}_{n=1}^{d_z}$ are fixed prior coefficients and are not trained. For each realization, the skew matrices used in the reversible operator reparameterization are
\begin{equation}
    \label{eq:n-genn_global_skew_matrices}
    \bm{A}_{i\bm{\vartheta}}(\mathbf{z})
    =
    \bm{B}_{i\bm{\vartheta}}(\mathbf{z})
    -
    \bm{B}_{i\bm{\vartheta}}(\mathbf{z})^{\top},
    \qquad i=1,\dots,d.
\end{equation}

In this construction, the state-dependent branch perturbs the reversible operator $\tilde{L}_{\bm{\psi}}(\mathbf{x})$ as a function of $(\mathbf{x},\mathbf{z})$, whereas the input-independent branch perturbs the global matrices $\bm{B}_{\bm{\psi}}$ as a function of $\mathbf{z}$ alone. Because both branches are driven by the same epistemic index, a single realization jointly perturbs the reversible operator $\tilde{L}_{\bm{\vartheta}}(\mathbf{x},\mathbf{z})$ and the skew-symmetric matrices $\{\bm{A}_{i\bm{\vartheta}}(\mathbf{z})\}_{i=1}^{d}$ used in the reparameterization. This preserves the state-independent character of $\bm{B}_{\bm{\psi}}$ while allowing the state-dependent uncertainty to be expressed through $\tilde{L}_{\bm{\vartheta}}$. Although introduced here for the global matrices in the GENERIC reparameterization, the input-independent branch can also be extended to global unknown quantities in inverse problems. In~\ref{sec:epinet_inverse_problems}, we further discuss the details, validate this extension on the Kraichnan--Orszag and Korteweg--de Vries benchmarks studied in~\cite{zou2024neuraluq}, and compare the proposed method with B-PINNs-HMC~\cite{yang2021b, zou2024neuraluq}.

\subsubsection{Structure-preserving reparameterization}
\label{sec:s-penn_reparameterization}
According to~\cite{osband2023epistemic}, the epinet framework does not prescribe a fixed architecture for the learnable and prior networks. This flexibility is central in the present structure-preserving setting, since the epinet attached to each thermodynamic building block can be selected to preserve the admissible function class required by the corresponding N-GENNs component. For the reversible contribution, the reversible operator epinet associated with $\tilde{L}_{\bm{\psi}}$ uses unconstrained neural networks for both the learnable network and the prior network, and incorporates the two-branch design of Section~\ref{sec:epinet_input_independent}. The energy and entropy epinets associated with $E_{\bm{\psi}}$ and $S_{\bm{\psi}}$ are also represented by unconstrained multilayer perceptrons (MLPs). 
Finally, the dissipation-potential epinet must preserve the 
convexity conditions imposed in N-GENNs. We therefore use the PICNNs for the raw dissipation-potential epinet to ensure the outputs are convex in $\mathbf{x}^{*}$.


Specifically, to enforce the reversible degeneracy condition, we apply the reversible operator reparameterization at each epistemic index. For sampled $\mathbf{z}$, the base networks in Eq.~\eqref{eq:n-genn_poisson_operator_reparam} are replaced by $S_{\bm{\vartheta}}$, $\tilde{L}_{\bm{\vartheta}}$, and the matrices defined in Eq.~\eqref{eq:n-genn_global_skew_matrices}. This yields the reparameterized reversible operator
\begin{equation}
    \label{eq:s-penn_poisson_reassembly}
    L_{\bm{\vartheta}}(\mathbf{x},\mathbf{z})
    =
    Q_{S_{\bm{\vartheta}}}(\mathbf{x},\mathbf{z})^{\top}
    \tilde{L}_{\bm{\vartheta}}(\mathbf{x},\mathbf{z})
    Q_{S_{\bm{\vartheta}}}(\mathbf{x},\mathbf{z}),
\end{equation}
where $Q_{S_{\bm{\vartheta}}}(\mathbf{x},\mathbf{z})$ is constructed from $\{\bm{A}_{i\bm{\vartheta}}(\mathbf{z})\}_{i=1}^{d}$ and $\mathrm{D}S_{\bm{\vartheta}}(\mathbf{x},\mathbf{z})$ in the same manner as in the deterministic construction.

For the irreversible contribution, we form the energy-orthogonal projection using the augmented energy,
\begin{equation}
    P_{E_{\bm{\vartheta}}}(\mathbf{x},\mathbf{z})
    =
    I-\frac{\mathrm{D} E_{\bm{\vartheta}}(\mathbf{x},\mathbf{z})\mathrm{D} E_{\bm{\vartheta}}(\mathbf{x},\mathbf{z})^{\top}}
    {\|\mathrm{D} E_{\bm{\vartheta}}(\mathbf{x},\mathbf{z})\|_2^2}.
\end{equation}
The augmented dissipation potential used in the S-PENNs dynamics is obtained by applying the same projection-based reparameterization to the augmented raw potential,
\begin{equation}
    \label{eq:s-penn_dissipation_reassembly}
    \begin{aligned}
    \Xi_{\bm{\vartheta}}(\mathbf{x},\mathbf{x}^{*},\mathbf{z})
    &=
    \tilde{\Xi}_{\bm{\vartheta}}\!\left(\mathbf{x},P_{E_{\bm{\vartheta}}}(\mathbf{x},\mathbf{z})\mathbf{x}^{*},\mathbf{z}\right)
    -
    \tilde{\Xi}_{\bm{\vartheta}}(\mathbf{x},\mathbf{0},\mathbf{z})
    -
    \left(
    \left.
        \mathrm{D}_{\mathbf{x}^{*}}\tilde{\Xi}_{\bm{\vartheta}}(\mathbf{x},\mathbf{x}^{*},\mathbf{z})
    \right|_{\mathbf{x}^{*}=\mathbf{0}}
    \right)^{\top}
    P_{E_{\bm{\vartheta}}}(\mathbf{x},\mathbf{z})\mathbf{x}^{*}.
    \end{aligned}
\end{equation}
Together with the constrained epinet parameterizations described above, each epistemic realization remains in the same admissible function classes as the deterministic N-GENNs building blocks.

It is worth noting that the dissipation-potential epinet preserves the required 
convexity only when their constrained components are combined with nonnegative coefficients. Since the epinet outputs in Eqs.~\eqref{eq:epinet_learnable_network}--\eqref{eq:epinet_prior_network} are weighted by the epistemic index, the coefficients multiplying the dissipation-potential components must be nonnegative. 
Since the reference distribution $\pi(\mathbf{z})$ can be freely chosen as stated in~\cite{osband2023epistemic}, $\pi(\mathbf{z})$ having nonnegative support is picked here. 

The constrained augmented building blocks define the S-PENNs vector field
\begin{equation}
    \label{eq:s-penn_dynamics}
    g_{\bm{\vartheta}}(\mathbf{x},\mathbf{z})
    =
    L_{\bm{\vartheta}}(\mathbf{x},\mathbf{z})\mathrm{D} E_{\bm{\vartheta}}(\mathbf{x},\mathbf{z})
    +
    \left.\mathrm{D}_{\mathbf{x}^{*}}\Xi_{\bm{\vartheta}}(\mathbf{x},\mathbf{x}^{*},\mathbf{z})\right|_{\mathbf{x}^{*}=\mathrm{D}S_{\bm{\vartheta}}(\mathbf{x},\mathbf{z})}.
\end{equation}
Since the reversible and irreversible constraints are built into the augmented parameterization, $g_{\bm{\vartheta}}(\cdot,\mathbf{z})$ is thermodynamically consistent for every fixed draw of $\mathbf{z}$.

\subsubsection{Training and predictive sampling}
\label{sec:spenn_training_prediction}
Let $\mathcal{D}_{\mathrm{train}}=\{\mathbf{X}_i\}_{i=1}^{N_{\mathrm{train}}}$ denote the training dataset, where $\mathbf{X}_i=(\mathbf{x}_i^{(t)})_{t=0}^{N_t}$ is the $i$-th reference rollout and $\mathbf{x}_i^{(0)}$ is its initial state. S-PENNs are trained in two stages. First, the deterministic base networks are fitted by minimizing
\begin{equation}
    \label{eq:base_training_loss}
    \mathcal{L}^{\mathrm{base}}_{\mathrm{S-PENNs}}(\bm{\psi})=\frac{1}{N_{\mathrm{train}}(N_t+1)}\sum_{i=1}^{N_{\mathrm{train}}}\sum_{t=0}^{N_t}
    \left\|g_{\bm{\psi}}(\mathbf{x}_i^{(t)})-\dot{\mathbf{x}}_i^{(t)}\right\|_2^2,
\end{equation}
where $\dot{\mathbf{x}}_i^{(t)}$ denotes the ground-truth time derivative. Second, the base-network parameters $\bm{\psi}$ are frozen and only the epinet parameters $\bm{\phi}$ are optimized. For each trajectory $i$, an epistemic index $\mathbf{z}_i\sim\pi$ is sampled and kept fixed over the rollout. The epinet objective is
\begin{equation}
    \label{eq:epinet_training_loss}
    \mathcal{L}^{\mathrm{epinet}}_{\mathrm{S-PENNs}}(\bm{\phi})
    =
    \mathbb{E}_{\mathbf{z}_i\overset{\mathrm{i.i.d.}}{\sim}\pi}
    \left[
    \frac{1}{N_{\mathrm{train}}(N_t+1)}
    \sum_{i=1}^{N_{\mathrm{train}}}\sum_{t=0}^{N_t}
    \left\|
        g_{\bm{\vartheta}}(\mathbf{x}_i^{(t)},\mathbf{z}_i)
        -
        \dot{\mathbf{x}}_i^{(t)}
    \right\|_2^2
    \right].
\end{equation}
At test time, let $N_s$ denote the number of predictive samples. 
For each testing trajectory $i$, independent draws $\{\mathbf{z}_i^{(r)}\}_{r=1}^{N_s}$ generate predictive rollouts $\{\hat{\mathbf{X}}_i^{(r)}\}_{r=1}^{N_s}$ initialized at $\mathbf{x}_i^{(0)}$. This ensemble is summarized by the empirical predictive mean $\hat{\mu}_{\bm{\vartheta}}$ and componentwise predictive standard deviation $\hat{\sigma}_{\bm{\vartheta}}$.

\subsection{Post-hoc calibration via split conformal prediction}
\label{sec:cp_calibration}
The sampled S-PENNs realizations are thermodynamically consistent by construction, but their empirical uncertainty summaries, the mean $\hat{\mu}_{\bm{\vartheta}}$ and componentwise standard deviation $\hat{\sigma}_{\bm{\vartheta}}$, do not by themselves guarantee to be well-calibrated. Standard regression calibration techniques, including variance scaling~\cite{levi2022evaluating} and CDF-based corrections~\cite{kuleshov2018accurate, zelikman2020crude}, are less suitable here because they need additional model fitting and can modify the implied predictive distribution. We instead use split conformal prediction~\cite{papadopoulos2002inductive, lei2018distribution} to calibrate the prediction intervals. This post-processing step leaves the sampled trajectories unchanged and gives distribution-free finite-sample marginal coverage under the exchangeability assumption stated next.

The calibration and testing samples used in the split conformal prediction are required to be jointly exchangeable~\cite{papadopoulos2002inductive, lei2018distribution}. For dynamical systems, the exchangeable unit is the full trajectory rather than an individual timestep, since time levels within one rollout are coupled by the evolution equations. Following trajectory-level conformal treatments for dynamical systems~\cite{stankeviciute2021conformal, sun2024copula, gopakumar2026uncertainty}, 
we regard each reference rollout $\mathbf{X}_i$ as one sample. Using the notation of Section~\ref{sec:spenn_training_prediction}, the available trajectories are partitioned into three disjoint datasets: a training dataset $\mathcal{D}_{\mathrm{train}}$ for training the S-PENNs (Section~\ref{sec:spenn_training_prediction}), a calibration dataset $\mathcal{D}_{\mathrm{cal}}$ of $N_{\mathrm{cal}}$ trajectories reserved for calibration, and a testing dataset $\mathcal{D}_{\mathrm{test}}$ used only for evaluation. In the numerical examples below, this assumption is satisfied by drawing the initial conditions independently from the same distribution for the calibration and testing datasets.

For each calibration trajectory $\mathbf{X}_i=(\mathbf{x}_i^{(t)})_{t=0}^{N_t}\in\mathcal{D}_{\mathrm{cal}}$, initialized at $\mathbf{x}_i^{(0)}$, let $j\in\mathcal{I}$ index one scalar entry of the corresponding forecast rollout. 
For ODE trajectories, $j$ specifies a state variable and a noninitial time point; for field-valued problems, it specifies a state variable and a noninitial space--time grid point. The initial state is prescribed as an input to the predictor and is therefore excluded from $\mathcal{I}$. Similarly, prescribed boundary conditions are also excluded from $\mathcal{I}$. Calibration is applied only to the forecast portion of the rollout, namely the time points $t=1,\dots,N_t$ for trajectory-valued problems and the corresponding non-prescribed space--time grid points for field-valued problems. 
We define the normalized nonconformity score~\cite{angelopoulos2023conformal}
\begin{equation}
    s_i(j)
    =
    \frac{\left|\mathbf{X}_i(j)-\hat{\mu}_{\bm{\vartheta}}(\mathbf{x}_i^{(0)};j)\right|}
    {\hat{\sigma}_{\bm{\vartheta}}(\mathbf{x}_i^{(0)};j)},
    \qquad
    i=1,\dots,N_{\mathrm{cal}}.
\end{equation}
Then, for each fixed index $j\in\mathcal{I}$, let
$s_{(1)}(j)\le\cdots\le s_{(N_{\mathrm{cal}})}(j)$
denote the ordered calibration scores at $j$, obtained by sorting the normalized nonconformity scores $\{s_i(j)\}_{i=1}^{N_{\mathrm{cal}}}$.
For a target miscoverage level $\alpha\in(0,1)$, set
\begin{equation}
    \label{eq:cp_quantile}
    r_\alpha
    =
    \left\lceil
        (N_{\mathrm{cal}}+1)(1-\alpha)
    \right\rceil,
    \qquad
    \hat{q}_{1-\alpha}(j)
    =
    \begin{cases}
        s_{(r_\alpha)}(j), & r_\alpha\le N_{\mathrm{cal}},\\
        +\infty, & r_\alpha=N_{\mathrm{cal}}+1.
    \end{cases}
\end{equation}
Here, $r_\alpha$ is the finite-sample-adjusted integer rank and
$\hat{q}_{1-\alpha}(j)$ is the corresponding split-conformal score
threshold at $j$. For a testing trajectory
$\mathbf{X}=(\mathbf{x}^{(t)})_{t=0}^{N_t}$, this threshold rescales
the predictive standard deviation to give
\begin{equation}
    \label{eq:cp_interval}
    \widehat{C}_{1-\alpha}(\mathbf{x}^{(0)};j)
    =
    \left[
    \hat{\mu}_{\bm{\vartheta}}(\mathbf{x}^{(0)};j)
    -
    \hat{q}_{1-\alpha}(j)
    \hat{\sigma}_{\bm{\vartheta}}(\mathbf{x}^{(0)};j),
    \;
    \hat{\mu}_{\bm{\vartheta}}(\mathbf{x}^{(0)};j)
    +
    \hat{q}_{1-\alpha}(j)
    \hat{\sigma}_{\bm{\vartheta}}(\mathbf{x}^{(0)};j)
    \right].
\end{equation}

For any testing trajectory that is jointly exchangeable with the calibration trajectories, the interval in Eq.~\eqref{eq:cp_interval} satisfies, for every fixed $j\in\mathcal{I}$,
\begin{equation}
    \label{eq:cp_coverage_guarantee}
    \mathbb{P}\!\left(
    \mathbf{X}(j)\in\widehat{C}_{1-\alpha}(\mathbf{x}^{(0)};j)
    \right)
    \ge 1-\alpha.
\end{equation}
This is the standard split-conformal rank argument~\cite{shafer2008tutorial, angelopoulos2023conformal}: exchangeability of the trajectories implies exchangeability of the $N_{\mathrm{cal}}$ calibration scores and the corresponding test score at the same index $j$. Therefore, the test score exceeds the empirical quantile $\hat{q}_{1-\alpha}(j)$ with probability at most $\alpha$. A proof of this result is given in Appendix~D of~\cite{angelopoulos2023conformal}, following~\cite{papadopoulos2002inductive}. This finite-sample statement is distribution-free in the sense that it does not assume a parametric data-generating law, a correctly specified predictive distribution, or a particular residual model. Its only statistical requirement is the exchangeability condition above.

\section{Numerical examples}
\label{sec:numerical_experiments}
In this section, we evaluate the proposed S-PENNs framework against two benchmark UQ methods, deep ensembles and MC dropout, on three numerical examples of increasing complexity. As discussed in Section~\ref{sec:s-penns}, the reference distribution for the epistemic index $\mathbf{z}$ is a modeling choice. We therefore consider three cases: a half-normal distribution, denoted ``S-PENNs (half-normal)''; a uniform distribution on $[0,5]$, denoted ``S-PENNs (uniform)''; and a unit-rate exponential distribution truncated to $[0,5]$, denoted ``S-PENNs (exponential)''. 
For quantitative evaluation, we consider the following metrics. Predictive accuracy for each state variable is measured by the relative $\ell^2$ error (RL2E) of the predictive mean over the simulated time interval or over the space--time grid, and by pointwise absolute errors for field-valued quantities. Probabilistic performance is evaluated with the continuous ranked probability score (CRPS) and the energy score (ES), both proper scoring rules for predictive distributions~\cite{gneiting2007strictly}. The CRPS is evaluated on scalar marginals and provides a componentwise measure expressed in the physical units of each predicted variable~\cite{matheson1976scoring, hersbach2000decomposition}. The ES extends this assessment to vector-valued probabilistic forecasts and is used here to evaluate the predictive distribution of the full rollout~\cite{gneiting2008assessing}. This is appropriate in the present setting because the learned dynamics couple the state variables, whereas scalar marginal scores do not assess their joint behavior. CRPS and ES are evaluated on the uncalibrated predictive samples from all methods. Meanwhile, the calibration quality of split conformal prediction is separately assessed through the empirical coverage of the calibrated S-PENNs prediction intervals. Unless otherwise stated, the calibrated prediction intervals shown in the trajectory and field plots use $\alpha=0.05$, corresponding to a target marginal coverage level of $1-\alpha=95\%$. Definitions of these metrics are given in~\ref{sec:eval_metric_define}.

\subsection{Harmonic oscillator in a heat bath}
\label{sec:harmonic_oscillator}
We first consider the harmonic oscillator example used in the N-GENNs study~\cite{votruba2026nonlinear}. This low-dimensional problem has a closed-form GENERIC representation, and its irreversible response is generated by a quadratic dissipation potential. It therefore provides a controlled setting for testing whether the proposed UQ construction can perturb a learned thermodynamic model without violating its conservation and dissipation properties.

The state of the system is given by $\mathbf{x}=(q,p,\epsilon)$, where $q$ and $p$ are the oscillator position and conjugate momentum, respectively, and $\epsilon$ represents the internal energy of the heat bath. The total energy and the canonical Poisson operator are given by
\begin{equation}
    E(\mathbf{x}) = \frac{p^2}{2m}+\frac{1}{2}kq^2+\epsilon,
    \qquad
    L =
    \begin{pmatrix}
        0 & 1 & 0 \\
        -1 & 0 & 0 \\
        0 & 0 & 0
    \end{pmatrix}.
\end{equation}
For a heat bath at constant temperature $T_{\mathrm{bath}}$, the entropy and the mobility matrix are
\begin{equation}
    S(\mathbf{x})=\frac{\epsilon}{T_{\mathrm{bath}}},
    \qquad
    M(\mathbf{x}) =
    \gamma T_{\mathrm{bath}}
    \begin{pmatrix}
        0 & 0 & 0 \\
        0 & 1 & -\frac{p}{m} \\
        0 & -\frac{p}{m} & \frac{p^2}{m^2}
    \end{pmatrix},
\end{equation}
where $m$ is the mass, $k$ is the stiffness of the linear spring connecting the mass to a fixed support, and $\gamma$ is the damping coefficient. The dissipation potential is
$\Xi(\mathbf{x},\mathbf{x}^{*})=\frac{1}{2}(\mathbf{x}^{*})^{T}M(\mathbf{x})\mathbf{x}^{*}$.
Altogether, these building blocks give the evolution equations
\begin{equation}
    \dot{q} = \frac{p}{m}, \qquad
    \dot{p} = -kq - \gamma\frac{p}{m}, \qquad
    \dot{\epsilon} = \gamma\frac{p^2}{m^2}.
\end{equation}

\subsubsection{Data generation and model validation}
For this example, the dataset is generated by using an implicit midpoint integrator for the ODE system above. The simulations use a uniform timestep $\Delta t=0.015$ and physical parameters $m=1.0$, $k=1.2$, and $\gamma=0.4$. Initial states are drawn uniformly from $q(0)\in[-1.0,\,1.0]$ and $p(0)\in[-1.0,\,1.0]$, while the heat-bath energy is initialized as $\epsilon(0)=0$. The generated dataset consists of $100$ trajectories with $1000$ state records per trajectory. The trajectory-level partition follows the protocol described in Section~\ref{sec:cp_calibration}: $60$ trajectories are used for training, $50$ for calibration, and $40$ for testing, where the splitting is performed randomly.

Model validation is performed by rolling out each trained model from the initial conditions of the testing trajectories and comparing the predictions with the corresponding simulated reference solutions. All rollout predictions use a standard fourth-order Runge--Kutta (RK4) integrator with the same timestep as the generated dataset, $\Delta t=0.015$. S-PENNs and MC dropout use $N_s=2000$ UQ realizations for each testing trajectory, while the deep ensembles use $50$ independently trained N-GENNs with different architecture choices and random initializations. The network architectures, optimization settings, and inference parameters are summarized in Table~\ref{tab:nn_hyperparameters} of~\ref{sec:nn_model_details}.

\subsubsection{Results and discussion}
\begin{figure}[htbp]
    \centering
    \includegraphics[width=\textwidth]{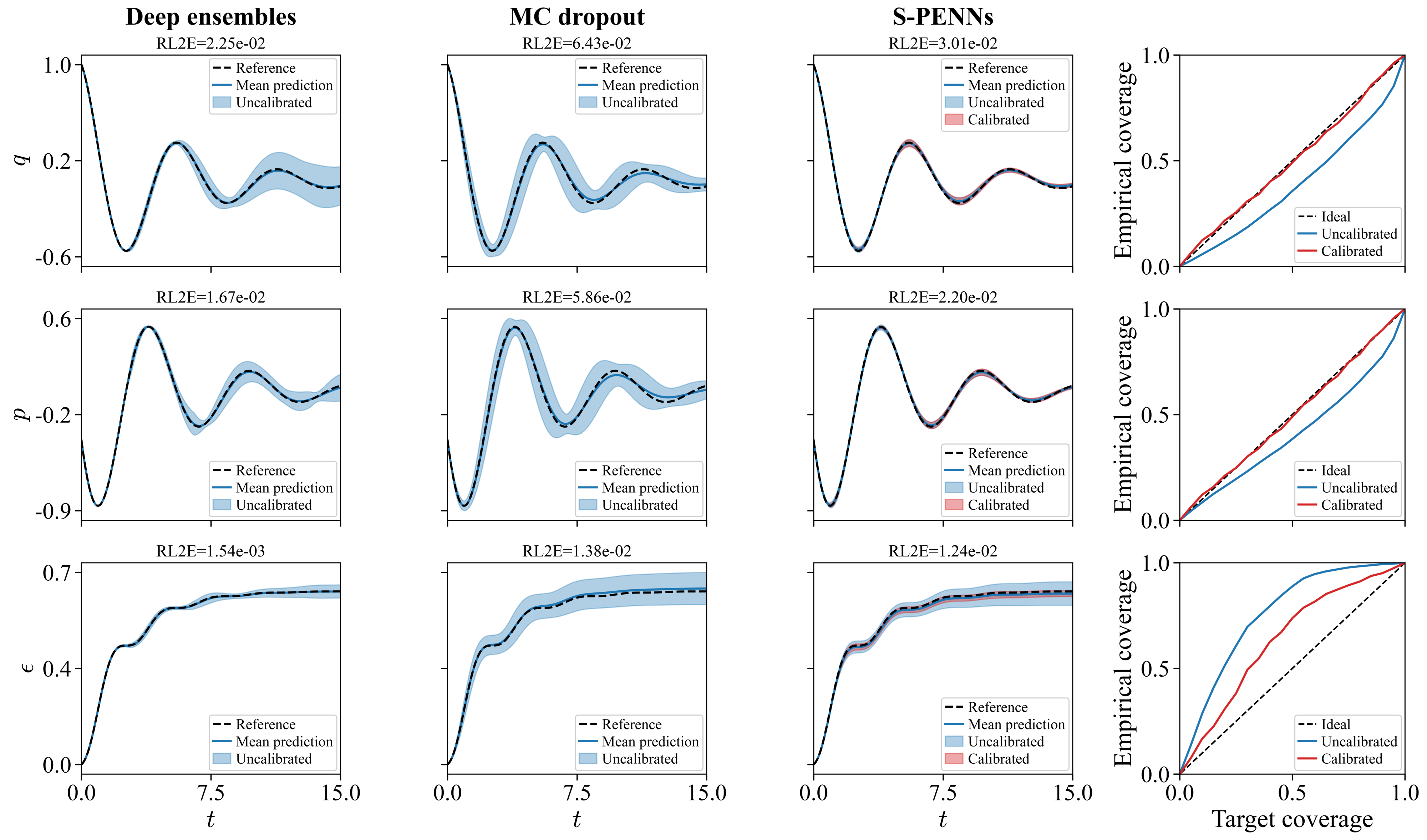}
    \caption{Predictive trajectories and calibration diagnostics for the harmonic oscillator example. The first three columns compare deep ensembles, MC dropout, and S-PENNs (half-normal) on a representative testing trajectory; rows correspond to $q$, $p$, and $\epsilon$. In each panel, the predictive mean is shown as a solid line, the reference trajectory as a dashed line, and the uncalibrated $\hat{\mu}\pm 2\hat{\sigma}$ band as a blue shaded region. For S-PENNs (half-normal), the calibrated 95\% conformal prediction interval is overlaid in red. The rightmost column plots empirical coverage against target coverage for S-PENNs (half-normal), with the diagonal indicating ideal calibration. Titles report the relative $\ell^2$ error (RL2E) for each method and state variable.}
    \label{fig:harmonic_main_results}
\end{figure}

Fig.~\ref{fig:harmonic_main_results} compares the predictions of deep ensembles, MC dropout, and S-PENNs (half-normal) on a representative testing trajectory. The first three columns correspond to the different methods, while the rows show the results for the state variables $q$, $p$, and $\epsilon$. Each panel displays the predictive trajectory together with its associated uncertainty band.
The rightmost column reports empirical coverage for the uncalibrated and calibrated S-PENNs (half-normal) intervals. All three methods reproduce the damped oscillator response, including the oscillatory exchange between $q$ and $p$ and the monotonic increase in the bath energy $\epsilon$. Deep ensembles give the smallest relative $\ell^2$ errors on this trajectory, with $2.25\times10^{-2}$ for $q$, $1.67\times10^{-2}$ for $p$, and $1.54\times10^{-3}$ for $\epsilon$. S-PENNs (half-normal) remain accurate on this displayed trajectory, with errors of $3.019\times10^{-2}$ for $q$, $2.20\times10^{-2}$ for $p$, and $1.24\times10^{-2}$ for $\epsilon$. MC dropout is the worst of the three methods for all three variables, giving errors of $6.43\times10^{-2}$ for $q$, $5.86\times10^{-2}$ for $p$, and $1.38\times10^{-2}$ for $\epsilon$. The uncertainty bands in the trajectory panels provide a qualitative comparison of predictive spread. MC dropout produces the broadest bands, whereas the other two methods produce narrower bands. The rightmost panels show that the uncalibrated S-PENNs deviate from the target coverage, as empirical coverage can lie above or below the target depending on the state variable and coverage level. Split conformal calibration moves the curves toward the ideal diagonal, improving coverage without changing the sampled trajectories.

\begin{figure}[htbp]
    \centering
    \includegraphics[width=\textwidth]{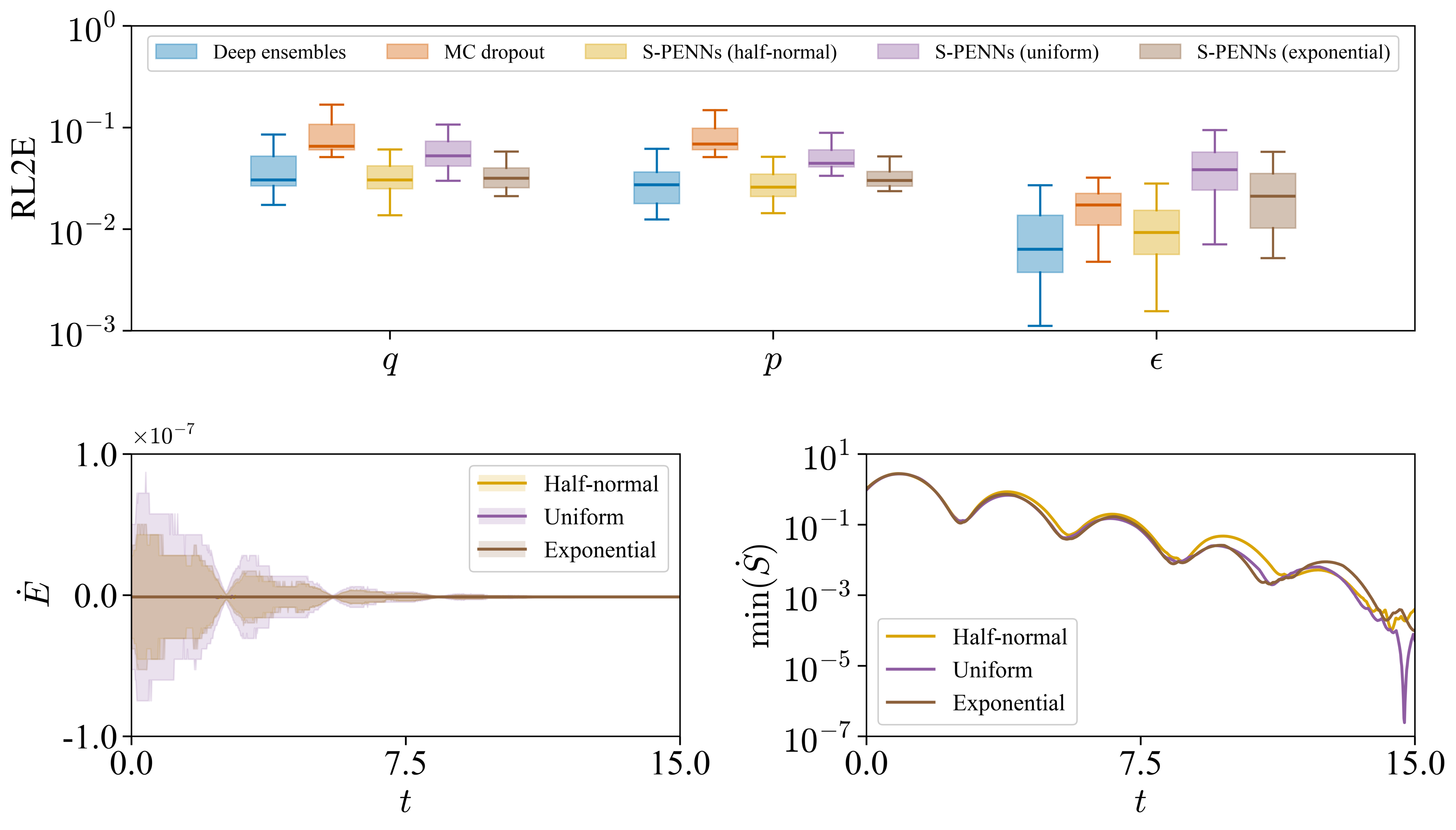}
    \caption{Predictive accuracy and structure-preservation diagnostics for the harmonic oscillator example. The top panel shows the componentwise trajectory-wise relative $\ell^2$ error (RL2E) over the testing dataset for $q$, $p$, and $\epsilon$, comparing deep ensembles, MC dropout, and S-PENNs using half-normal, uniform, and exponential distributions for the epistemic index. The box plots show the median with the solid lines and the boxes span the interquartile ranges. The bottom-left panel shows the energy rate $\dot{E}$ over time, with the median as a solid line and the 95\% prediction interval as a shaded band across sampled rollouts. The bottom-right panel shows the pointwise minimum entropy production $\min(\dot{S})$ over time on a logarithmic scale, where the minimum is taken across the sampled S-PENNs rollouts at each timestep.}
    \label{fig:harmonic_combined_diagnostics}
\end{figure}

As a broader assessment, Fig.~\ref{fig:harmonic_combined_diagnostics} extends the comparison to the full testing dataset and examines the thermodynamic structure of the S-PENNs rollouts. The top panel reports the trajectory-wise relative $\ell^2$ errors. All three S-PENNs variants have lower median errors than MC dropout for $q$ and $p$. Among them, the half-normal distribution gives the lowest median errors for all three state variables. Its medians for $q$ and $p$ are nearly identical to those of deep ensembles, while deep ensembles retain the lowest median error for $\epsilon$. The uniform and exponential variants of S-PENNs have higher median errors on $\epsilon$ than MC dropout. The bottom panels assess the first and second laws for the sampled rollouts of the representative testing trajectory. For all three S-PENNs variants, the median energy rate remains centered at zero, and the 95\% intervals stay at the $10^{-7}$ scale. The pointwise minimum entropy-production rate over the sampled rollouts remains positive throughout the trajectory and decreases overall as the damped oscillator approaches equilibrium. These results indicate that, in this example, S-PENNs preserve thermodynamic consistency across all three reference distributions, while predictive accuracy remains stable, with slightly greater variation for $\epsilon$ than for $q$ and $p$.

Beyond predictive accuracy and thermodynamic consistency, computational cost and the statistical quality of the resulting uncertainty estimates are central to the practical value of the UQ method. 
Table~\ref{tab:harmonic_cost_and_scores} compares serial wall time and proper scoring rules computed from the uncalibrated predictive samples. Deep ensembles attain the lowest ES and CRPS values, while all S-PENNs variants outperform MC dropout in ES and in CRPS for $q$ and $p$; the half-normal variant also improves CRPS for $\epsilon$. Measured wall times are $9.1\times10^{3}$~s for the 50-member deep ensemble, $2.3\times10^{2}$~s for MC dropout, and $1.2\times10^{2}$--$1.3\times10^{2}$~s for S-PENNs. Thus, S-PENNs are less costly despite using 2000 rather than 50 samples. Scaling the deep ensemble to 2000 members gives $3.6\times10^{5}$~s, which is three orders of magnitude higher than the cost for S-PENNs.

\begin{table}[htbp]
    \caption{Serial wall time and proper scoring rules for the harmonic oscillator example. The wall time includes model training and generation of predictive samples for all testing trajectories. ES denotes the energy score over the full multivariate trajectory, and CRPS is averaged for each state variable. Scores are computed from uncalibrated predictive samples and averaged over the testing dataset. Lower ES and CRPS values indicate better probabilistic predictions.}
    \label{tab:harmonic_cost_and_scores}
    \centering
    \small
    \begin{tabular}{lccccc}
        \toprule
        Method & Wall time (sec) & ES $\downarrow$ & CRPS($q$) $\downarrow$ & CRPS($p$) $\downarrow$ & CRPS($\epsilon$) $\downarrow$ \\
        \midrule
        Deep ensembles$^{\ast}$ & $9.1 \times 10^{3}$ & $6.8 \times 10^{-2}$ & $1.1 \times 10^{-3}$ & $9.3 \times 10^{-4}$ & $5.1 \times 10^{-4}$ \\
        MC dropout & $2.3 \times 10^{2}$ & $7.0 \times 10^{-1}$ & $1.2 \times 10^{-2}$ & $1.3 \times 10^{-2}$ & $4.5 \times 10^{-3}$ \\
        S-PENNs (half-normal) & $1.2 \times 10^{2}$ & $2.4 \times 10^{-1}$ & $4.1 \times 10^{-3}$ & $3.8 \times 10^{-3}$ & $3.0 \times 10^{-3}$ \\
        S-PENNs (uniform) & $1.3 \times 10^{2}$ & $4.9 \times 10^{-1}$ & $7.4 \times 10^{-3}$ & $7.4 \times 10^{-3}$ & $7.8 \times 10^{-3}$ \\
        S-PENNs (exponential) & $1.3 \times 10^{2}$ & $3.2 \times 10^{-1}$ & $4.2 \times 10^{-3}$ & $4.3 \times 10^{-3}$ & $5.9 \times 10^{-3}$ \\
        \bottomrule
    \end{tabular}
    \vspace{2pt}
    \parbox{0.95\linewidth}{\footnotesize $^{\ast}$ For deep ensembles, the reported wall time is the serial cost of the 50-member ensemble, and the scoring rules are computed from the same ensemble.}
\end{table}

Overall, the harmonic oscillator example shows that S-PENNs can introduce epistemic uncertainty into a hard-constrained learned dynamics model at a low computational cost while guaranteeing thermodynamic consistency. Across the testing dataset, S-PENNs produce accurate mean predictions, improve clearly over MC dropout, and remain largely insensitive to the choice of reference distribution. The sampled rollouts preserve energy conservation and nonnegative entropy production, and split conformal prediction improves empirical coverage without modifying these admissible samples. Although deep ensembles remain the strongest baseline in accuracy and proper scoring rules for this low-dimensional problem, S-PENNs achieve competitive predictive accuracy and UQ performance at substantially lower computational cost.

\subsection{Idealized chemical motor}
\label{sec:chemical_motor}
The second example is the idealized chemical motor considered in the N-GENNs study~\cite{votruba2026nonlinear}. A piston of mass $m$ and cross-sectional area $A$ is attached to a linear spring with natural length $q_0$ and encloses a van der Waals mixture undergoing a chemical reaction  $\mathrm{X}_1+\mathrm{X}_2\rightleftharpoons\mathrm{X}_3$. The absolute piston position $q$ determines the chamber volume through $V(q)=Aq$, while $q-q_0$ is the spring extension. The state variables and vector of species mole numbers are
\begin{equation}
    \mathbf{x}=(q,p,\epsilon,n_1,n_2,n_3),\qquad
    \bm{n}=(n_1,n_2,n_3)^T,
\end{equation}
where $p$ is the piston momentum, $\epsilon$ is the internal energy of the gas mixture, and $n_\mu$ is the number of moles of specie $\mathrm{X}_\mu$. The total energy combines the kinetic energy of the piston, the elastic energy of the spring, and the internal energy of the mixture,
\begin{equation}
    E(\mathbf{x})=\frac{p^2}{2m}+\frac{1}{2}k(q-q_0)^2+\epsilon,
    \label{eq:chemical_motor_energy}
\end{equation}
where $k$ is the spring stiffness. 
The Poisson operator for this system is noncanonical and couples the piston momentum to the gas internal energy as,
\begin{equation}
    L(\mathbf{x})=
    \begin{pmatrix}
        0 & 1 & 0 & 0 & 0 & 0 \\
        -1 & 0 & F(\mathbf{x}) & 0 & 0 & 0 \\
        0 & -F(\mathbf{x}) & 0 & 0 & 0 & 0 \\
        0 & 0 & 0 & 0 & 0 & 0 \\
        0 & 0 & 0 & 0 & 0 & 0 \\
        0 & 0 & 0 & 0 & 0 & 0
    \end{pmatrix},\qquad
    F(\mathbf{x})=
    \frac{(\partial S/\partial q)_{p,\epsilon}}
         {(\partial S/\partial\epsilon)_{p,q}}.
    \label{eq:chemical_motor_poisson}
\end{equation}
Here $F(\mathbf{x})$ is the force exerted by the gas on the piston.

We adopt the N-GENNs dimensionless convention $R=h=N_A=1$, where $R$, $h$, and $N_A$ denote the gas constant, Planck constant, and Avogadro constant, respectively. With $S=S_{\mathrm{phys}}/R$, the gas-constant-scaled van der Waals entropy is
\begin{equation}
    S(\mathbf{x})=
    \sum_{\mu=1}^{3} n_\mu
    \left[
        \frac{5}{2}+\ln\left(
        \frac{V-\sum_{\nu=1}^{3}n_\nu b_\nu}{n_\mu}
        \left[
            \frac{4\pi m_\mu}{3\sum_{\beta=1}^{3}n_\beta}
            \left(
                \epsilon+
                \frac{\sum_{\rho,\sigma=1}^{3}
                n_\rho n_\sigma a_{\rho\sigma}}{V}
            \right)
        \right]^{3/2}
        \right)
    \right],
    \label{eq:chemical_motor_entropy}
\end{equation}
here $m_\mu$ is the molecular mass of species $\mathrm{X}_\mu$, $b_\mu$ is its excluded-volume parameter, and $a_{\rho\sigma}$ is the interaction parameter between species $\mathrm{X}_\rho$ and $\mathrm{X}_\sigma$. The mass-action kinetics are generated by the non-quadratic dissipation potential
\begin{equation}
    \Xi(\mathbf{x},\mathbf{x}^{*})=
    \alpha\sqrt{n_1n_2n_3}
    \left[
        \cosh\left(\frac{n_1^{*}+n_2^{*}-n_3^{*}}{2}\right)-1
    \right],
    \label{eq:chemical_motor_dissipation}
\end{equation}
where $\mathbf{x}^{*}=(q^{*},p^{*},\epsilon^{*},n_1^{*},n_2^{*},n_3^{*})$ is the conjugate state, and $\alpha>0$ is the rate coefficient. Evaluating the GENERIC evolution with Eqs.~\eqref{eq:chemical_motor_energy}--\eqref{eq:chemical_motor_dissipation} gives
\begin{equation}
    \begin{aligned}
        \dot q={}&\frac{p}{m},\\
        \dot p={}&-k(q-q_0)
        +\frac{2}{3}
        \frac{\epsilon Aq+\sum_{\rho,\sigma=1}^{3}
        n_\rho n_\sigma a_{\rho\sigma}}
        {q\left(Aq-\sum_{\nu=1}^{3}n_\nu b_\nu\right)}
        -\frac{\sum_{\rho,\sigma=1}^{3}
        n_\rho n_\sigma a_{\rho\sigma}}{Aq^2},\\
        \dot\epsilon={}&-\frac{p}{m}
        \left[
        \frac{2}{3}
        \frac{\epsilon Aq+\sum_{\rho,\sigma=1}^{3}
        n_\rho n_\sigma a_{\rho\sigma}}
        {q\left(Aq-\sum_{\nu=1}^{3}n_\nu b_\nu\right)}
        -\frac{\sum_{\rho,\sigma=1}^{3}
        n_\rho n_\sigma a_{\rho\sigma}}{Aq^2}
        \right],\\
        \dot n_1={}&\frac{\alpha}{4}
        \left(Kn_3-\frac{n_1n_2}{K}\right),\\
        \dot n_2={}&\frac{\alpha}{4}
        \left(Kn_3-\frac{n_1n_2}{K}\right),\\
        \dot n_3={}&\frac{\alpha}{4}
        \left(\frac{n_1n_2}{K}-Kn_3\right).
    \end{aligned}
    \label{eq:chemical_motor_dynamics}
\end{equation}
Here $K$ is the reaction constant. With $n=\sum_{\beta=1}^{3}n_\beta$, $\bm{a}=(a_{\rho\sigma})$, and $\bm{b}=(b_1,b_2,b_3)^T$, it is given by
\begin{equation}
    K=
    \exp\left[
        \frac{3n\sum_{\sigma=1}^{3}(a_{1\sigma}+a_{2\sigma}-a_{3\sigma})n_\sigma}
        {2(\epsilon Aq+\bm{n}^{T}\bm{a}\bm{n})}
        -\frac{n(b_1+b_2-b_3)}
        {2(Aq-\bm{n}^{T}\bm{b})}
    \right]
    \left(\frac{m_1m_2}{m_3}\right)^{3/4}
    (Aq-\bm{n}^{T}\bm{b})^{1/2}
    \left[
        \frac{4\pi}{3n}
        \left(
            \epsilon+\frac{\bm{n}^{T}\bm{a}\bm{n}}{Aq}
        \right)
    \right]^{3/4}.
    \label{eq:chemical_motor_reaction_constant}
\end{equation}
More details on this example can be found in Section 4.2.1 and Appendix B of~\cite{votruba2026nonlinear}.

\subsubsection{Data generation and model validation}
For this example, the dataset is generated using the implicit midpoint integrator. The simulations use a uniform timestep $\Delta t=0.01$, and the physical parameters
\begin{equation}
    \begin{gathered}
        m=0.8,\quad A=1.0,\quad q_0=2.0,\quad k=31.6,\quad \alpha=2.0,\\
        \bm{a}=
        \begin{pmatrix}
            0.1 & 0.1 & 0.1\\
            0.1 & 0.1 & 0.1\\
            0.1 & 0.1 & 20.0
        \end{pmatrix},\quad
        \bm{b}=(0.05,0.05,0.05)^{T},\quad
        (m_1,m_2,m_3)=(1.0,1.2,2.2).
    \end{gathered}
\end{equation}
The piston starts from rest, $p(0)=0$, while the initial mole numbers, piston position, and dimensionless temperature $T_0$ are sampled uniformly and independently from
\begin{equation}
    \begin{gathered}
        q(0)\in[1.2,2.8],\qquad T_0\in[0.8,1.2],\\
        n_1(0)\in[0.8,1.3],\qquad
        n_2(0)\in[1.2,2.4],\qquad
        n_3(0)\in[0.002,0.003].
    \end{gathered}
\end{equation}
The initial internal energy is set to $\epsilon(0)=\tfrac{3}{2}n(0)T_0$, which gives $\epsilon(0)\in[2.4,6.7]$ to the reported precision. The generated dataset consists of 220 trajectories with 1000 state records per trajectory on $t\in[0,9.99]$. Among these generated trajectories, 140 trajectories are used for training, 50 for calibration, and 30 for testing.

Model validation is performed by rolling out each trained model from the initial condition of every testing trajectory and comparing the predictions with the corresponding reference solution. RK4 integrator is used with the same timestep employed to generate the dataset, $\Delta t=0.01$. For each test trajectory, S-PENNs and MC dropout use $N_s=2000$ uncertainty realizations, whereas the deep-ensemble method comprises 50 independently trained N-GENNs with different architecture choices and random initializations. Further details on the network architectures, optimization settings, and inference parameters are provided in Table~\ref{tab:nn_hyperparameters}.

\subsubsection{Results and discussion}
\begin{figure}[htbp]
    \centering
    \includegraphics[width=\textwidth]{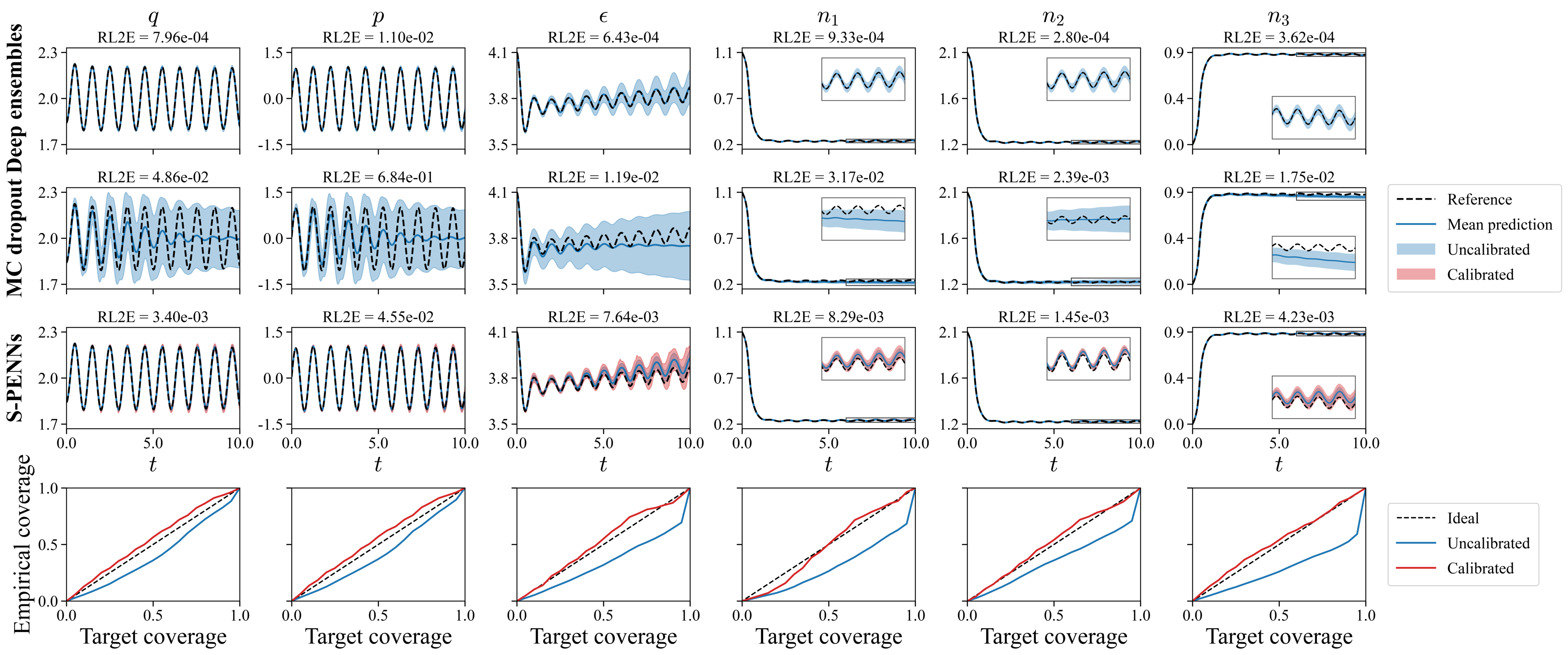}
    \caption{Predictive trajectories and calibration diagnostics for the idealized chemical motor. Columns correspond to $q$, $p$, $\epsilon$, $n_1$, $n_2$, and $n_3$. The first three rows show deep ensembles, MC dropout, and S-PENNs (half-normal), respectively, on a representative testing trajectory. The predictive mean is plotted as a solid line, the simulated reference as a dashed line, and the uncalibrated $\hat{\mu}\pm2\hat{\sigma}$ interval as a blue band. The S-PENNs row also includes the calibrated 95\% conformal interval in red. Insets enlarge the late-time response of the three species. The final row compares empirical and target coverage for the uncalibrated and calibrated S-PENNs intervals. Titles report the componentwise relative $\ell^2$ error (RL2E).}
    \label{fig:chemical_motor_main_results}
\end{figure}

In Fig.~\ref{fig:chemical_motor_main_results}, we compare the three UQ methods on a representative testing trajectory. The six columns correspond to the state variables $q$, $p$, $\epsilon$, $n_1$, $n_2$, and $n_3$, while the first three rows show deep ensembles, MC dropout, and S-PENNs (half-normal), respectively. Deep ensembles produce the most accurate predictive mean for every component, with relative $\ell^2$ error of $7.96\times10^{-4}$ for $q$, $1.10\times10^{-2}$ for $p$, $6.43\times10^{-4}$ for $\epsilon$, $9.33\times10^{-4}$ for $n_1$, $2.80\times10^{-4}$ for $n_2$, and $3.62\times10^{-4}$ for $n_3$. S-PENNs (half-normal) gives corresponding errors of $3.40\times10^{-3}$, $4.55\times10^{-2}$, $7.64\times10^{-3}$, $8.29\times10^{-3}$, $1.45\times10^{-3}$, and $4.23\times10^{-3}$. For both methods, the momentum $p$ has the largest error because small phase discrepancies are amplified as the trajectory repeatedly changes sign. MC dropout shows larger phase and amplitude deviations, with errors of $4.86\times10^{-2}$ for $q$, $6.84\times10^{-1}$ for $p$, $1.19\times10^{-2}$ for $\epsilon$, $3.17\times10^{-2}$ for $n_1$, $2.39\times10^{-3}$ for $n_2$, and $1.75\times10^{-2}$ for $n_3$. Additionally, the presented uncertainty bands provide a further distinction between the methods on this representative testing trajectory. The bands obtained from deep ensembles remain narrow and successfully cover the errors. MC dropout gives the widest bands, particularly for $q$ and $p$. However, this larger spread does not fully cover the discrepancies in $n_1$ and $n_3$. The uncalibrated S-PENNs bands are comparable to those of deep ensembles. The coverage curves in the bottom row of Fig.~\ref{fig:chemical_motor_main_results} confirm that these raw bands systematically under-cover, indicating the model is overconfident. Split conformal calibration successfully brings every component close to the target without altering the underlying thermodynamically admissible samples.

\begin{figure}[htbp]
    \centering
    \includegraphics[width=\textwidth]{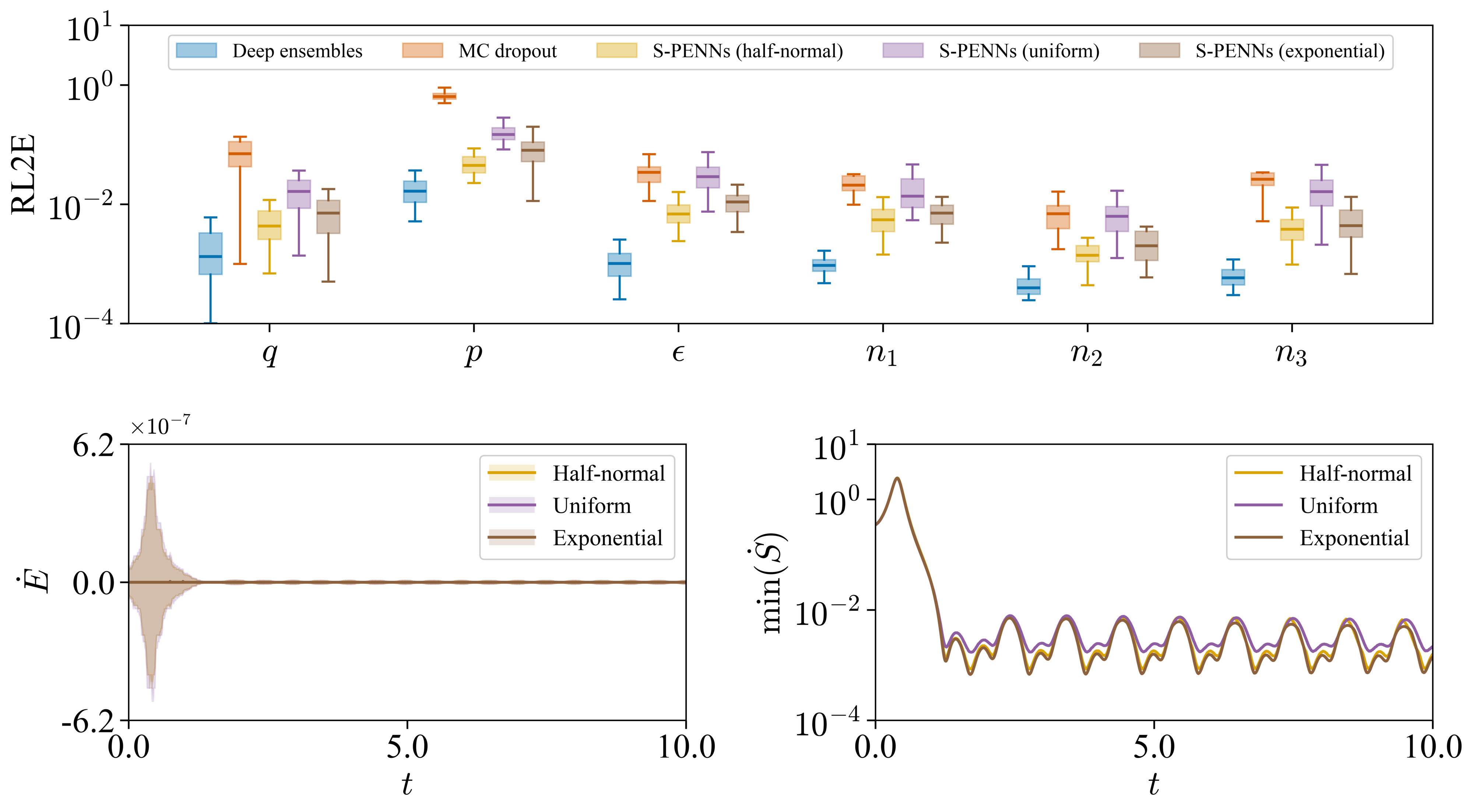}
    \caption{Predictive accuracy and structure-preservation diagnostics for the idealized chemical motor. The top panel shows the componentwise trajectory-wise relative $\ell^2$ error (RL2E) over the testing dataset for $q$, $p$, $\epsilon$, $n_1$, $n_2$, and $n_3$, comparing deep ensembles, MC dropout, and S-PENNs using half-normal, uniform, and exponential distributions for the epistemic index. The box plots show the medians as solid lines, while the boxes span the interquartile ranges. The bottom-left panel shows the energy rate $\dot{E}$ over time for a representative testing trajectory, with the median as a solid line and the 95\% prediction interval as a shaded band across sampled S-PENNs rollouts. The bottom-right panel shows the pointwise minimum entropy-production rate $\min(\dot{S})$ on a logarithmic scale, where the minimum is taken across the sampled S-PENNs rollouts at each timestep.}
    \label{fig:chemical_motor_combined_diagnostics}
\end{figure}
The comparison over the full test dataset is shown in Fig.~\ref{fig:chemical_motor_combined_diagnostics}. Deep ensembles have the lowest median relative $\ell^2$ error for every component. MC dropout constantly produces the largest errors, while all S-PENNs variants have a lower median than MC dropout across all six variables. The lower panels verify the thermodynamic behavior of the S-PENNs samples. For all three reference distributions, the energy-rate median remains at zero and the 95\% interval stays on the order of $10^{-7}$. The minimum sampled entropy-production rate remains positive throughout the rollout. Thus, as intended by the framework’s design, the uncertainty realizations retain the first- and second-law structure even in the presence of nonlinear chemical kinetics and state-dependent reversible coupling.

\begin{table}[htbp]
    \caption{Serial wall time and proper scoring rules for the idealized chemical motor example. The wall time includes model training and generation of predictive samples for all testing trajectories. ES denotes the energy score over the full multivariate trajectory, and CRPS is averaged for each state variable. Scores are computed from uncalibrated predictive samples and averaged over the testing dataset. Lower ES and CRPS values indicate better probabilistic predictions.}
    \label{tab:chemical_motor_cost_and_scores}
    \centering
    \scriptsize
    \setlength{\tabcolsep}{3pt}
    \resizebox{\textwidth}{!}{%
    \begin{tabular}{lcccccccc}
        \toprule
        Method & Wall time (sec) & ES $\downarrow$ & CRPS($q$) $\downarrow$ & CRPS($p$) $\downarrow$ & CRPS($\epsilon$) $\downarrow$ & CRPS($n_1$) $\downarrow$ & CRPS($n_2$) $\downarrow$ & CRPS($n_3$) $\downarrow$ \\
        \midrule
        Deep ensembles$^{\ast}$ & $1.5\times10^{4}$ & $8.2\times10^{-1}$ & $3.5\times10^{-3}$ & $1.8\times10^{-2}$ & $8.1\times10^{-3}$ & $6.0\times10^{-4}$ & $6.3\times10^{-4}$ & $6.8\times10^{-4}$ \\
        MC dropout & $4.4\times10^{2}$ & $1.6\times10^{1}$ & $7.4\times10^{-2}$ & $3.7\times10^{-1}$ & $8.7\times10^{-2}$ & $6.4\times10^{-3}$ & $4.7\times10^{-3}$ & $1.3\times10^{-2}$ \\
        S-PENNs (half-normal) & $3.2\times10^{2}$ & $2.0\times10^{0}$ & $7.5\times10^{-3}$ & $3.8\times10^{-2}$ & $2.5\times10^{-2}$ & $2.0\times10^{-3}$ & $1.6\times10^{-3}$ & $1.8\times10^{-3}$ \\
        S-PENNs (uniform) & $2.7\times10^{2}$ & $6.0\times10^{0}$ & $2.1\times10^{-2}$ & $1.0\times10^{-1}$ & $8.6\times10^{-2}$ & $5.7\times10^{-3}$ & $5.6\times10^{-3}$ & $7.1\times10^{-3}$ \\
        S-PENNs (exponential) & $3.1\times10^{2}$ & $2.7\times10^{0}$ & $1.1\times10^{-2}$ & $5.4\times10^{-2}$ & $3.1\times10^{-2}$ & $2.3\times10^{-3}$ & $2.0\times10^{-3}$ & $2.2\times10^{-3}$ \\
        \bottomrule
    \end{tabular}%
    }
    \vspace{2pt}
    \parbox{0.95\linewidth}{\footnotesize $^{\ast}$ For deep ensembles, the reported wall time is the serial cost of the 50-member ensemble, and the scoring rules are computed from the same ensemble.}
\end{table}

To complete the comparison, Table~\ref{tab:chemical_motor_cost_and_scores} reports probabilistic scores and serial computational time. Deep ensembles give the lowest ES, $8.2\times10^{-1}$, and the lowest componentwise CRPS values. Measured wall times are $1.5\times10^{4}$~s for the 50-member deep ensemble, $4.4\times10^{2}$~s for MC dropout, and $2.7\times10^{2}$--$3.2\times10^{2}$~s for S-PENNs. Thus, S-PENNs are about two times less costly compared to deep ensembles and outperform MC dropout in probabilistic scores. Scaling the deep ensemble to 2000 members gives $5.9\times10^{5}$~s, which is about three orders of magnitude higher than the cost needed for S-PENNs. These results extend the proposed construction to systems with noncanonical reversible coupling and a non-quadratic reaction potential.

\subsection{Viscoplastic model}
\label{sec:viscoplastic_model}
The third example is designed to assess S-PENNs on a field-valued problem 
characterized by a nonsmooth nonlinear dissipation law, moving beyond finite-dimensional ODE dynamics, as considered in the first two examples. More specifically, we consider a one-dimensional continuum system governed by a viscoplastic constitutive relation of Perzyna type ~\cite{lubliner2008plasticity,simo1998computational}, whose governing equations have been put in GENERIC form in~\cite{mielke2011formulation}. The state vector  $\mathbf{x}=(u,p,\varepsilon^{vp},\theta)$, consists of the displacement field $u(x,t)$, momentum field $p(x,t)$, viscoplastic strain field $\varepsilon^{vp}(x,t)$, and temperature field $\theta(x,t)$. 
The total and elastic strains are
\begin{equation}
    \varepsilon=\frac{\partial u}{\partial x},\qquad
    \varepsilon^e=\varepsilon-\varepsilon^{vp}.
\end{equation}
Neglecting heat conduction, the energy and the entropy functionals are
\begin{equation}
    E[\mathbf{x}] = \int \left(\frac{1}{2\rho}p^2 + c\theta + \frac{1}{2}C(\varepsilon - \varepsilon^{vp})^2\right)\,\mathrm{d}x,
    \qquad
    S[\mathbf{x}] = \int c\log\theta\,\mathrm{d}x,
\end{equation}
where $\rho$, $C$, and $c$ denote the mass density, elastic modulus, and specific heat, respectively. The reversible part is governed by the canonical Poisson operator
\begin{equation}
    L = \begin{pmatrix}
        0 & I & 0 & 0 \\
        -I & 0 & 0 & 0 \\
        0 & 0 & 0 & 0 \\
        0 & 0 & 0 & 0
    \end{pmatrix},
\end{equation}
whereas the irreversible response is governed by a non-quadratic Perzyna-type dissipation potential
\begin{equation}
    \Xi[\mathbf{x};\mathbf{x}^{*}] = \int \frac{\theta}{2\eta}\left\langle\left|\xi_{\varepsilon^{vp}} + \frac{\xi_\theta}{c}\sigma\right| - \frac{\sigma_y}{\theta}\right\rangle^2\,\mathrm{d}x,
\end{equation}
where $\mathbf{x}^{*}=(\xi_u,\xi_p,\xi_{\varepsilon^{vp}},\xi_\theta)$ denotes the dual variables, $\eta$ is the viscosity, $\sigma=C(\varepsilon-\varepsilon^{vp})$ is the stress, $\sigma_y>0$ is the constant yield stress, and $\langle a\rangle=\max(a,0)$ is the Macaulay bracket. The potential is convex with respect to the dual variables and non-quadratic and non-smooth due to the overstress threshold. Substitution into the GENERIC evolution equation gives
\begin{equation}
    \dot{u} = \frac{p}{\rho},\qquad
    \dot{p} = \frac{\partial}{\partial x}\left(C(\varepsilon - \varepsilon^{vp})\right),\qquad
    \dot{\varepsilon}^{vp} = \frac{1}{\eta}\langle|\sigma|-\sigma_y\rangle\,\mathrm{sign}(\sigma),\qquad
    \dot{\theta} = \frac{\sigma\dot{\varepsilon}^{vp}}{c}.
\end{equation}

\subsubsection{Data generation and model validation}
The dataset is generated by solving the evolution equations above on a bar of length $l=0.2$~m over $t\in[0,2.0\times10^{-4}]$~s. The material parameters are
\begin{equation}
    \rho = 7800\;\mathrm{kg\cdot m^{-3}},\quad C = 210\;\mathrm{GPa},\quad \eta = 8\;\mathrm{GPa\cdot s},\quad \sigma_y = 250\;\mathrm{MPa},\quad c = 1000\;\mathrm{J\cdot kg^{-1}\cdot K^{-1}}.
\end{equation}
All simulations start from $u(x,0)=0$, $p(x,0)=0$, $\varepsilon^{vp}(x,0)=0$, and $\theta(x,0)=298$~K. A uniform mesh with $N_x=29$ cells gives $\Delta x=6.90\times10^{-3}$~m, with displacement and momentum stored at nodes, and strain, viscoplastic strain, and temperature stored at cell centers. A Courant--Friedrichs--Lewy (CFL) number of $0.5$ based on the elastic wave speed $\sqrt{C/\rho}$ gives $\Delta t=6.65\times10^{-7}$~s. The nodal mechanical variables are advanced with velocity--Verlet, and the cell-centered viscoplastic strain and temperature are advanced with a forward Euler scheme, as in \citep{votruba2026nonlinear}. 

The left boundary is fixed, $u(0,t)=0$, while the right boundary follows a smooth ramp to a target engineering strain $\varepsilon_t$ and is then held fixed as
\begin{equation}
    u(l,t) = \begin{cases}
        \varepsilon_t\, l\left(3\left(\frac{t}{t_{\mathrm{ramp}}}\right)^2 - 2\left(\frac{t}{t_{\mathrm{ramp}}}\right)^3\right), & t \le t_{\mathrm{ramp}},\\
        \varepsilon_t\, l, & \text{otherwise},
    \end{cases}
\end{equation}
where $t_{\mathrm{ramp}}=1.6\times10^{-4}$~s. The dataset contains $150$ loading cases with uniformly spaced target strains $\varepsilon_t\in[1\%,2\%]$. As in the harmonic oscillator example, the split uses $70$ loading cases for training, $50$ for conformal calibration, and $30$ for testing.

The deterministic backbone follows the N-GENNs parameterization reviewed in Section~\ref{sec:N-GENNs_intro}. Because the state variables are fields, the networks learn the local energy, entropy, and dissipation densities rather than the corresponding functionals, which are obtained by integrating these densities over the domain. The kinetic contribution to the energy density, $p^2/(2\rho)$, is imposed analytically, while the remaining energy density, and the total entropy and dissipation densities are learned from data. The Poisson operator is fixed to the canonical form above, consistent with the GENERIC structure for generalized standard materials~\cite{mielke2011formulation}; hence, S-PENNs perturb only the density networks in this example. During rollout, $\dot{u}=p/\rho$ and the prescribed boundary histories are imposed directly, while the learned model supplies the interior momentum rate and the cell-centered rates of $\varepsilon^{vp}$ and $\theta$.

For evaluation, the learned model is fed with the initial conditions of each testing case and rolled out on the same one-dimensional mesh using an RK4 scheme with the same timestep used to generate the dataset, $\Delta t=6.65\times10^{-7}$~s. The rollout state consists of the interior nodal values of $u$ and $p$ together with all cell-centered values of $\varepsilon^{vp}$ and $\theta$. Boundary histories for $u$ and $p$ are imposed from the corresponding simulated case. S-PENNs and MC dropout use $N_s=2000$ stochastic rollouts per testing case, while the deep-ensemble baseline uses $50$ independently initialized N-GENNs with the width choices reported in Table~\ref{tab:nn_hyperparameters}. The full architecture, optimization, and inference settings are summarized in~\ref{sec:nn_model_details}.

\subsubsection{Results and discussion}
\begin{figure}[htbp]
    \centering
    \includegraphics[width=\textwidth]{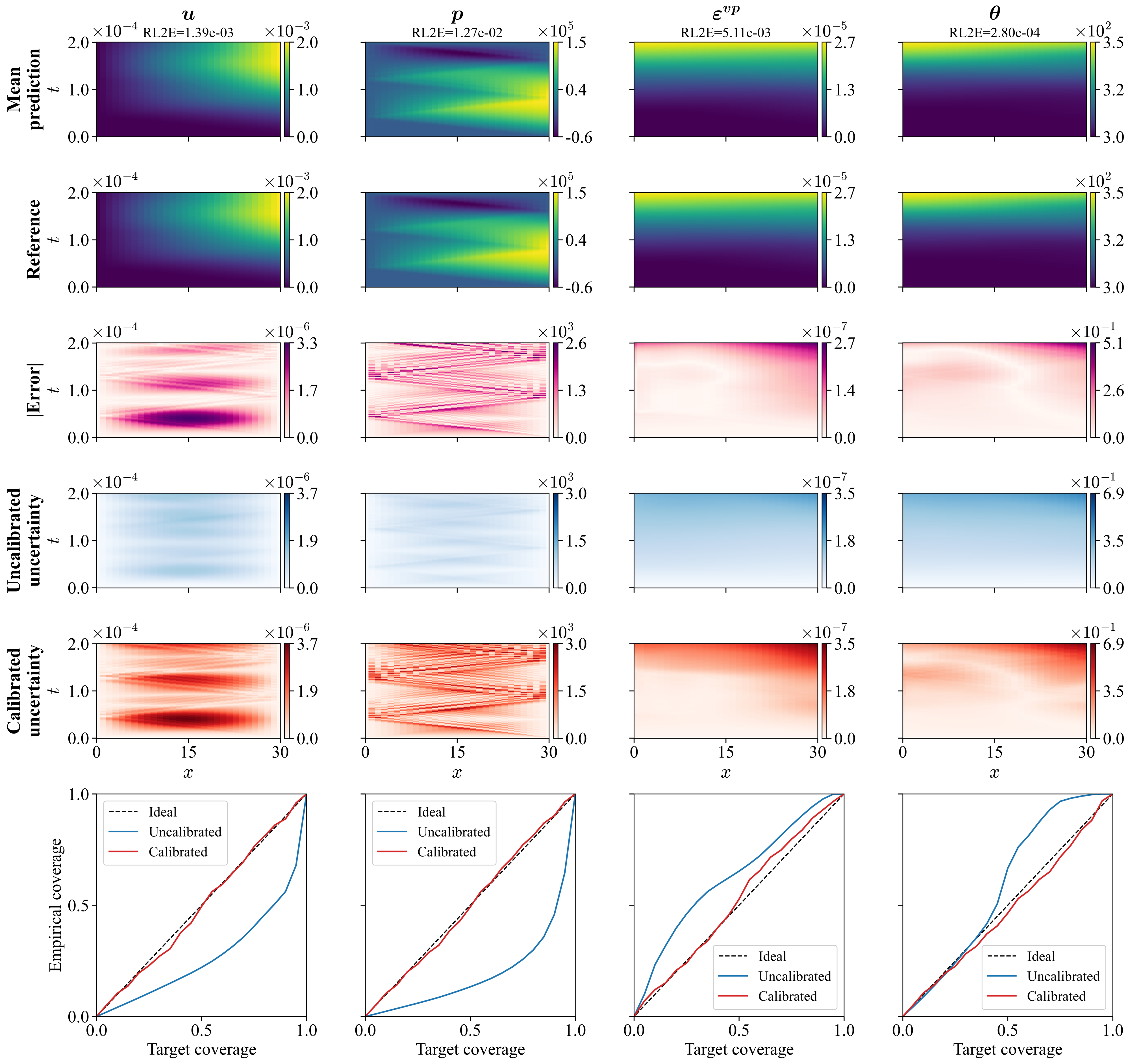}
    \caption{S-PENNs (half-normal) prediction, uncertainty, and calibration fields for the 1D viscoplastic model on a representative testing case. Columns correspond to $u$, $p$, $\varepsilon^{vp}$, and $\theta$. From top to bottom, the rows show the predictive mean, the reference field, the pointwise absolute error, the uncalibrated uncertainty given by the pointwise empirical standard deviation across sampled rollouts, the calibrated uncertainty given by the conformal interval half-width $\hat{q}_{1-\alpha}\hat{\sigma}$, and empirical coverage against target coverage. Titles report the relative $\ell^2$ error (RL2E) for each state variable.}
    \label{fig:viscoplastic_main_results}
\end{figure}

We present S-PENNs (half-normal) on a representative testing case for this example in Fig.~\ref{fig:viscoplastic_main_results}, with the corresponding deep ensemble and MC dropout baseline results provided in~\ref{sec:viscoplastic_baseline_results}. In these field visualizations, columns correspond to $u$, $p$, $\varepsilon^{vp}$, and $\theta$, and the rows show their predictive mean, simulated reference field, pointwise absolute error, and uncalibrated uncertainty. Fig.~\ref{fig:viscoplastic_main_results} further reports 
the calibrated uncertainty and the empirical coverage against target coverage for S-PENNs (half-normal). For this testing case, S-PENNs (half-normal) yield relative $\ell^2$ errors of $1.39\times10^{-3}$ for $u$, $1.27\times10^{-2}$ for $p$, $5.11\times10^{-3}$ for $\varepsilon^{vp}$, and $2.80\times10^{-4}$ for $\theta$. The corresponding errors for deep ensembles are $1.85\times10^{-3}$, $1.82\times10^{-2}$, $1.72\times10^{-3}$, and $1.30\times10^{-4}$, respectively. MC dropout produces the largest errors overall with $5.13\times10^{-3}$ for $u$, $5.09\times10^{-2}$ for $p$, $9.99\times10^{-3}$ for $\varepsilon^{vp}$, and $1.63\times10^{-3}$ for $\theta$. Despite these quantitative differences, none of the methods qualitatively reproduces the spatial patterns in the pointwise absolute error fields. The last row of Fig.~\ref{fig:viscoplastic_main_results} compares the empirical and target coverage of S-PENNs (half-normal). Before calibration, the uncertainty estimates under-cover for $u$ and $p$ but over-cover for $\varepsilon^{vp}$ and $\theta$, indicating overconfidence in the former two variables and underconfidence in the latter two. After split conformal calibration, the four coverage curves follow the target diagonal more closely. After split conformal calibration, all four empirical coverage curves follow the target diagonal more closely. The calibrated uncertainty fields also provide a closer representation of both the spatial error patterns and the range of error magnitudes for each state variable. These findings show that conformal prediction calibration improves coverage reliability in this PDE example.


\begin{figure}[htbp]
    \centering
    \includegraphics[width=1.0\textwidth]{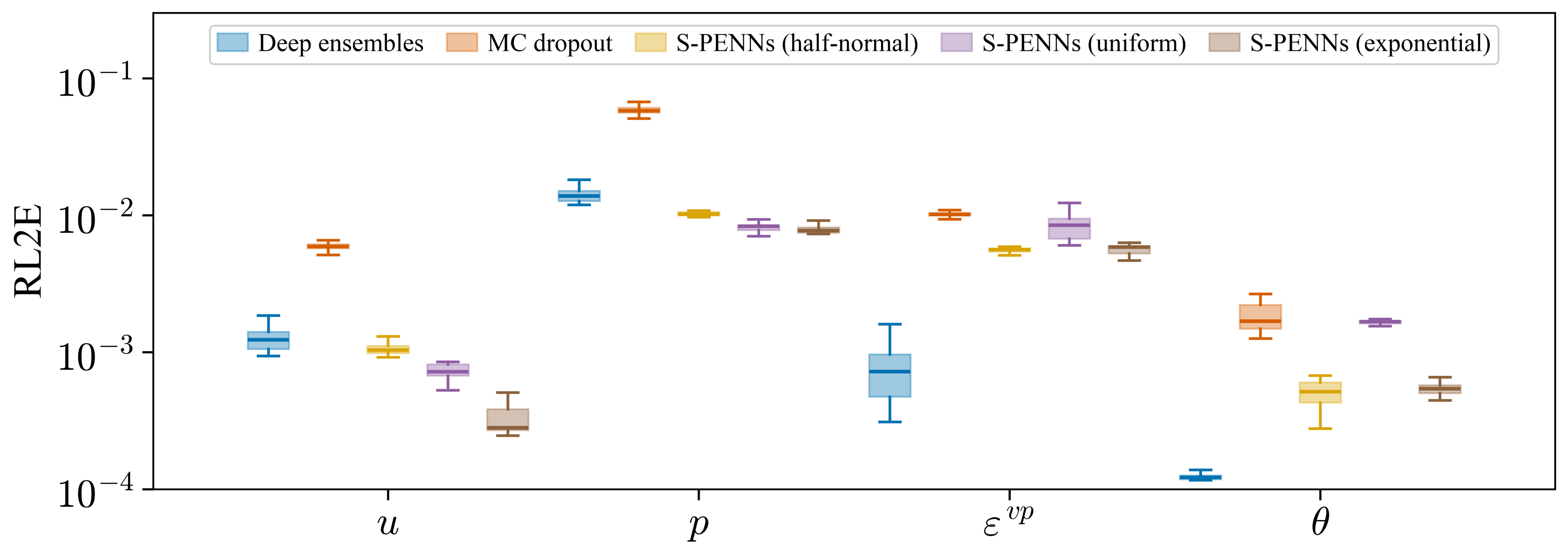}
    \caption{Componentwise testing-case relative $\ell^2$ error (RL2E) over the space--time grid for the 1D viscoplastic testing dataset. The grouped boxplots compare deep ensembles, MC dropout, and the half-normal, uniform, and exponential S-PENNs variants for $u$, $p$, $\varepsilon^{vp}$, and $\theta$. Each box shows the median as a solid line and boxes span the interquartile ranges.}
    \label{fig:viscoplastic_rl2e_boxplot}
\end{figure}

Fig.~\ref{fig:viscoplastic_rl2e_boxplot} moves from the representative case to the full testing dataset by plotting the statistics of the componentwise relative $\ell^2$ error for each testing case. 
MC dropout has the largest median RL2E for every field. The deep ensemble has the lowest median errors for $\varepsilon^{vp}$ and $\theta$, while S-PENNs (exponential) have the lowest median errors for $u$ and $p$. Among the S-PENNs variants, half-normal gives the lowest median errors for $\varepsilon^{vp}$ and $\theta$, whereas uniform and exponential reduce the $u$ and $p$ errors relative to half-normal. Thus, no S-PENNs reference distribution dominates across all four variables, but all three variants have lower median errors than MC dropout for every component. 

To compare the quality of the quantified uncertainty and the computational cost, Table~\ref{tab:viscoplastic_cost_and_scores} reports ES and componentwise CRPS from the uncalibrated samples together with serial wall time. All three S-PENNs variants achieve lower ES values than the deep ensemble, ranging from $4.8\times10^{4}$ to $6.6\times10^{4}$, compared with $1.0\times10^{5}$. MC dropout performs worst, with an ES of $3.0\times10^{5}$. The componentwise CRPS results show that S-PENNs (exponential) achieves the lowest values for $u$ and $p$, whereas the deep ensemble performs best for $\varepsilon^{vp}$ and $\theta$. Each S-PENNs variant also improves all four componentwise CRPS values relative to MC dropout while requiring less than half its wall time. Specifically, the S-PENNs runs take $1.0\times10^{3}$~s, compared with $2.3\times10^{3}$~s for MC dropout. Measured wall times is $7.5\times10^{4}$~s for the 50-member deep ensemble, which is about one order of magnitude higher than the cost for S-PENNs. Scaling the deep ensemble to 2000 members gives $3.0\times10^{6}$~s, or approximately $3.0\times10^{3}$ times the S-PENNs cost.

\begin{table}[htbp]
    \caption{Serial wall time and proper scoring rules for the 1D viscoplastic model. The wall time includes model training and generation of predictive samples for all testing cases. For deep ensembles, the reported time is extrapolated from the measured 50-member ensemble as described in the table note. ES denotes the energy score over the full-state rollout, and CRPS is reported for each physical field. Scores are computed from uncalibrated predictive samples and averaged over the testing dataset. Lower ES and CRPS values indicate better probabilistic predictions.}
    \label{tab:viscoplastic_cost_and_scores}
    \centering
    \small
    \begin{tabular}{lcccccc}
        \toprule
        Method & Wall time (sec) & ES $\downarrow$ & CRPS($u$) $\downarrow$ & CRPS($p$) $\downarrow$ & CRPS($\varepsilon^{vp}$) $\downarrow$ & CRPS($\theta$) $\downarrow$ \\
        \midrule
        Deep ensembles$^{\ast}$ & $7.5 \times 10^{4}$ & $1.0 \times 10^{5}$ & $1.1 \times 10^{-6}$ & $9.0 \times 10^{2}$ & $7.5 \times 10^{-9}$ & $1.9 \times 10^{-2}$ \\
        MC dropout & $2.3 \times 10^{3}$ & $3.0 \times 10^{5}$ & $3.3 \times 10^{-6}$ & $2.4 \times 10^{3}$ & $1.2 \times 10^{-7}$ & $3.6 \times 10^{-1}$ \\
        S-PENNs (half-normal) & $1.0 \times 10^{3}$ & $6.6 \times 10^{4}$ & $7.4 \times 10^{-7}$ & $5.7 \times 10^{2}$ & $3.8 \times 10^{-8}$ & $7.1 \times 10^{-2}$ \\
        S-PENNs (uniform) & $1.0 \times 10^{3}$ & $4.8 \times 10^{4}$ & $4.8 \times 10^{-7}$ & $3.9 \times 10^{2}$ & $8.5 \times 10^{-8}$ & $2.2 \times 10^{-1}$ \\
        S-PENNs (exponential) & $1.0 \times 10^{3}$ & $4.8 \times 10^{4}$ & $2.4 \times 10^{-7}$ & $3.8 \times 10^{2}$ & $3.8 \times 10^{-8}$ & $1.1 \times 10^{-1}$ \\
        \bottomrule
    \end{tabular}

    \vspace{2pt}
    \parbox{0.95\linewidth}{\footnotesize $^{\ast}$ For deep ensembles, the reported wall time is the serial cost of the 50-member ensemble, and the scoring rules are computed from the same ensemble.}
\end{table}

Taken together, the 1D viscoplastic results extend the ODE findings from the previous examples to a PDE setting where uncertainty must be quantified over spatiotemporal fields rather than finite-dimensional trajectories. All three S-PENNs variants achieve consistently strong predictive accuracy and comparable overall performance, while outperforming MC dropout in both predictive accuracy and probabilistic scores. They also provide competitive UQ performance relative to the deep ensemble, with a much lower computational cost.
These results show that the advantages observed in the ODE example carry over to field-valued PDE dynamics.

\section{Conclusion}
\label{sec:conclusion}
This work introduces Structure-Preserving Epistemic Neural Networks (S-PENNs) for uncertainty quantification in scientific machine learning models with architecturally enforced physical constraints. In the GENERIC setting, S-PENNs attach block-specific epinets to the thermodynamic building components of N-GENNs, which serve as base networks. The epinets share a common epistemic index. 
The resulting augmented blocks are then assembled through the same structure-preserving reparameterizations as in N-GENNs. As a result, each sampled vector field remains thermodynamically consistent by construction, preserving total energy and nonnegative entropy production at the level of individual realizations. Split conformal prediction is then used to calibrate interval widths obtained from the ensemble mean and standard deviation, providing finite-sample marginal coverage under trajectory-level exchangeability. The numerical examples show that S-PENNs provide a practical balance between structural fidelity, predictive quality, and computational cost across both finite-dimensional and field-valued dynamical systems. S-PENNs required typically $1-3$ orders of magnitude less serial wall time than deep ensembles, improved substantially over MC dropout in probabilistic scores, and showed only moderate sensitivity to the reference distribution for the epistemic index.

The present study is limited in two main respects. First, all three numerical examples use noise-free state measurements. With noisy states, the problem becomes an errors-in-variables (EiV) problem~\cite{fuller2009measurement, carroll2006measurement,soderstrom2007errors,zou2025uncertainty}: the clean trajectory should satisfy energy conservation and nonnegative entropy production, but the 
noise-corrupted trajectory need not. This differs from intrinsic physical stochasticity, for which individual realizations are governed by the appropriate balance and dissipation laws. Training the same constrained dynamics directly on noisy measurements would therefore impose the thermodynamic constraints on a noise-contaminated trajectory, which can bias the learned dynamics. Recent work in scientific machine learning has begun to address such EiV problems, for example in~\cite{zou2025uncertainty}. Second, epinets are designed as lightweight auxiliary networks for UQ~\cite{osband2023epistemic} and have been shown to match large ensembles at orders-of-magnitude lower computational cost in benchmarks~\cite{osband2023approximate}. However, the present study evaluates S-PENNs only on two low-dimensional ODE systems and a one-dimensional PDE system, and the computational cost advantage of S-PENNs relative to deep ensembles is expected to be further amplified in higher-dimensional settings.

Future work will first address noisy state measurements by separating the thermodynamically admissible trajectory from the measurement process. One route is to denoise the measurements before learning the constrained dynamics, while another is to use latent-variable formulations that represent noisy inputs before dynamics identification~\cite{conti2026veni}. A second direction is to evaluate scalability on higher-dimensional thermomechanical systems, where the relative cost of epinet sampling, rollout generation, and deep ensembles may differ from the present benchmarks. 

More broadly, the S-PENNs principle is not tied to GENERIC dynamics. It is relevant to other structure-preserving or physics-constrained machine learning models in computational mechanics and may also be useful for pretrained foundation models in science and engineering whose architectures or adaptation procedures encode physical priors. In these settings, uncertainty estimates should remain compatible with the encoded structure. Because epinets can be attached to frozen or partially frozen building blocks, they provide a lightweight route to physically admissible uncertainty samples for active learning, material design, and reliability assessment.

\renewcommand{\appendixname}{Appendix}
\appendix
\section{Evaluation metrics for the numerical examples}
\label{sec:eval_metric_define}
This appendix defines the evaluation metrics used in Section~\ref{sec:numerical_experiments}. To distinguish the testing data from the training trajectories in Section~\ref{sec:spenn_training_prediction}, let $i=1,\dots,N_{\mathrm{test}}$ index a trajectory in $\mathcal{D}_{\mathrm{test}}$, let $\mathbf{X}_i$ denote its reference rollout, and let $\hat{\mathbf{X}}_i^{(r)}$, $r=1,\dots,N_s$, denote its $r$-th predictive rollout sample. The scalar index~$j$ identifies a state variable at a discrete time point and, for a field-valued variable, at a spatio-temporal grid point. For each reported variable~$\nu$, $\mathcal{J}_{\nu}$ is the set of retained forecast entries used by the componentwise metric: forecast time points for an ODE variable and retained space--time grid points for a field-valued variable. The predictive sample mean at entry~$j$ is $\hat{\mu}_i(j)=N_s^{-1}\sum_{r=1}^{N_s}\hat{\mathbf{X}}_i^{(r)}(j)$. All metrics use the forecast portion of each rollout and therefore exclude the prescribed initial state at $t=0$, and prescribed boundary conditions, if applicable. Thus, the harmonic-oscillator and chemical-motor metrics use $t=1,\dots,N_t$. In the viscoplastic example, the imposed boundary conditions of $u$ and $p$ are also excluded.

\paragraph{Relative $\ell^2$ error}
Predictive accuracy is measured by the relative $\ell^2$ error (RL2E) of the predictive mean. For testing trajectory~$i$ and state variable~$\nu$,
\begin{equation}
    \mathrm{RL2E}_{i}^{(\nu)}
    =
    \left(
    \frac{
    \sum_{j\in\mathcal{J}_{\nu}}
    \left(\hat{\mu}_{i}(j)-\mathbf{X}_i(j)\right)^2
    }{
    \sum_{j\in\mathcal{J}_{\nu}}
    \left(\mathbf{X}_i(j)\right)^2
    }
    \right)^{1/2}.
\end{equation}
Figure titles report $\mathrm{RL2E}_{i}^{(\nu)}$ for the displayed trajectory, and the box plots summarize its distribution over the testing dataset. For field-valued variables, the figures also show the pointwise absolute error $|\hat{\mu}_i(j)-\mathbf{X}_i(j)|$.

\paragraph{Empirical coverage}
For a nominal coverage level $1-\alpha$, let $\widehat{C}_{1-\alpha,i}(j)$ denote the prediction interval at entry~$j$ of testing trajectory~$i$. For the uncalibrated rollout samples, its endpoints are the empirical $\alpha/2$ and $1-\alpha/2$ quantiles; for calibrated results, $\widehat{C}_{1-\alpha,i}(j)=\widehat{C}_{1-\alpha}(\mathbf{x}_i^{(0)};j)$ is the split-conformal interval in Eq.~\eqref{eq:cp_interval}. Empirical coverage is the fraction of reference rollout entries contained in these intervals~\cite{angelopoulos2023conformal}:
\begin{equation}
    \mathrm{EC}^{(\nu)}(1-\alpha)
    =
    \frac{1}{N_{\mathrm{test}}|\mathcal{J}_{\nu}|}
    \sum_{i=1}^{N_{\mathrm{test}}}
    \sum_{j\in\mathcal{J}_{\nu}}
    \mathbf{1}\!\left\{
    \mathbf{X}_i(j)
    \in
    \widehat{C}_{1-\alpha,i}(j)
    \right\},
\end{equation}
where $\mathbf{1}\{\cdot\}$ is the indicator function. 
The set $\mathcal{J}_{\nu}$ contains only forecast entries, namely, it excludes the prescribed initial conditions and boundary conditions.
Ideal empirical calibration corresponds to $\mathrm{EC}^{(\nu)}(1-\alpha)=1-\alpha$. Curves below this diagonal indicate under-coverage associated with overconfident models, whereas curves above it indicate over-coverage associated with underconfident models~\cite{gneiting2007probabilistic, angelopoulos2023conformal, xu2021conformal}. We remark that finite calibration and testing datasets can produce deviations from the diagonal even when the conformal procedure satisfies its marginal coverage guarantee~\cite{angelopoulos2023conformal}.

\paragraph{Proper scoring rules}
Probabilistic forecast quality is evaluated on the uncalibrated rollout samples using the continuous ranked probability score (CRPS) and the energy score (ES). These proper scoring rules 
assess calibration and sharpness jointly, with sharpness considered subject to calibration~\cite{gneiting2007strictly, gneiting2007probabilistic}. Lower values indicate better probabilistic forecasts.
The CRPS is computed on scalar marginals and averaged separately for each state variable~\cite{matheson1976scoring, hersbach2000decomposition}:
\begin{equation}
    \mathrm{CRPS}^{(\nu)}
    =
    \frac{1}{N_{\mathrm{test}}|\mathcal{J}_{\nu}|}
    \sum_{i=1}^{N_{\mathrm{test}}}
    \sum_{j\in\mathcal{J}_{\nu}}
    \left[
    \frac{1}{N_s}\sum_{r=1}^{N_s}
    \left|\hat{\mathbf{X}}_{i}^{(r)}(j)-\mathbf{X}_i(j)\right|
    -
    \frac{1}{2N_s^2}
    \sum_{r,r'=1}^{N_s}
    \left|\hat{\mathbf{X}}_{i}^{(r)}(j)-\hat{\mathbf{X}}_{i}^{(r')}(j)\right|
    \right].
\end{equation}
The ES extends the same assessment to the joint predictive distribution of the full rollout~\cite{gneiting2008assessing}. Viewing all forecast entries that remain after the exclusions above as vectors $\mathbf{X}_i$ and $\hat{\mathbf{X}}_i^{(r)}$, respectively, we estimate
\begin{equation}
    \mathrm{ES}
    =
    \frac{1}{N_{\mathrm{test}}}
    \sum_{i=1}^{N_{\mathrm{test}}}
    \left[
    \frac{1}{N_s}\sum_{r=1}^{N_s}
    \left\|\hat{\mathbf{X}}_{i}^{(r)}-\mathbf{X}_i\right\|_2
    -
    \frac{1}{2N_s^2}
    \sum_{r,r'=1}^{N_s}
    \left\|\hat{\mathbf{X}}_{i}^{(r)}-\hat{\mathbf{X}}_{i}^{(r')}\right\|_2
    \right].
\end{equation}

\section{Additional baseline results}
\label{sec:viscoplastic_baseline_results}
\begin{figure}[htbp]
    \centering
    \includegraphics[width=\textwidth]{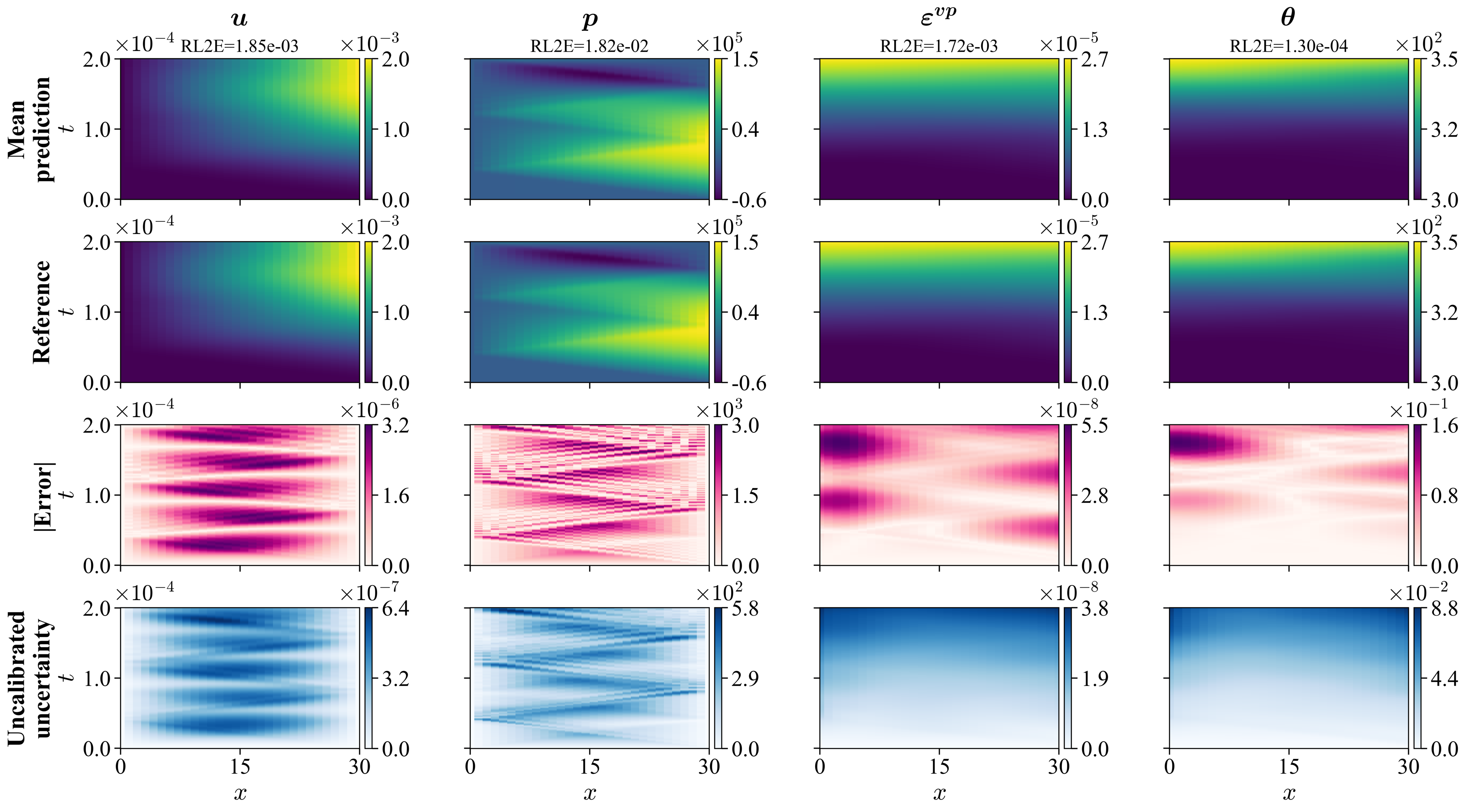}
    \caption{Deep ensemble prediction and uncertainty fields for the 1D viscoplastic model on a representative testing instance. Columns correspond to $u$, $p$, $\varepsilon^{vp}$, and $\theta$. From top to bottom, the rows show the predictive mean, the simulated solution, the pointwise absolute error, and the uncalibrated uncertainty given by the pointwise empirical standard deviation across ensemble members. Titles report the relative $\ell^2$ error (RL2E) for each state variable.}
    \label{fig:viscoplastic_ensembles_results}
\end{figure}

\begin{figure}[htbp]
    \centering
    \includegraphics[width=\textwidth]{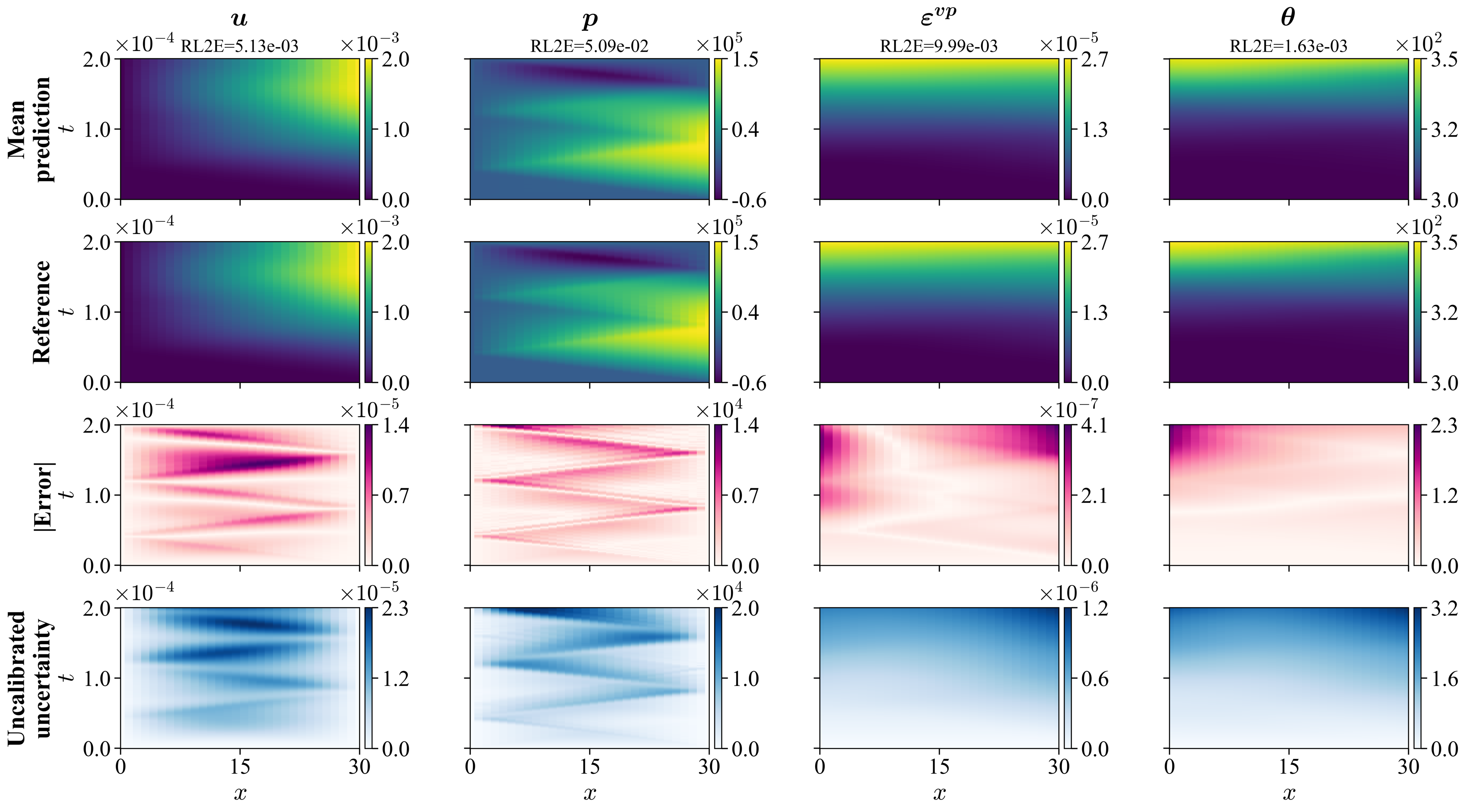}
    \caption{MC dropout prediction and uncertainty fields for the 1D viscoplastic model on a representative testing instance. Columns correspond to $u$, $p$, $\varepsilon^{vp}$, and $\theta$. From top to bottom, the rows show the predictive mean, the simulated solution, the pointwise absolute error, and the uncalibrated uncertainty given by the pointwise empirical standard deviation across stochastic forward passes. Titles report the relative $\ell^2$ error (RL2E) for each state variable.}
    \label{fig:viscoplastic_dropout_results}
\end{figure}

This appendix reports the representative field visualizations on the same testing case for the deep ensembles and MC dropout baselines referenced in Section~\ref{sec:viscoplastic_model}. 
Figures~\ref{fig:viscoplastic_ensembles_results} and~\ref{fig:viscoplastic_dropout_results} use the same layout as the S-PENNs results in Fig.~\ref{fig:viscoplastic_main_results}. The four columns correspond to $u$, $p$, $\varepsilon^{vp}$, and $\theta$, and the rows display, from top to bottom, the predictive mean, simulated reference field, pointwise absolute error, and uncalibrated uncertainty. The uncertainty is the pointwise empirical standard deviation across the 50 independently trained models for the deep ensemble and across 2000 stochastic rollouts for MC dropout.

\section{Neural-network setting, training, and inference details}
\label{sec:nn_model_details}
All models are implemented in Python with JAX and Flax. Training and inference are performed in single precision on a single NVIDIA RTX A6000 GPU. Unless otherwise stated, the N-GENNs backbone architectures and base-network training settings used by each UQ method follow the settings for the corresponding numerical examples in the original N-GENNs paper~\cite{votruba2026nonlinear}. Table~\ref{tab:nn_hyperparameters} summarizes the architecture, optimization, and inference settings used in the harmonic oscillator, idealized chemical motor, and 1D viscoplastic model. The following paragraphs detail method-specific implementation choices not fully captured by the table.

\paragraph{S-PENNs}
S-PENNs' training follows the two-stage procedure described in Section~\ref{sec:spenn_training_prediction}: the deterministic N-GENNs backbone is trained first, and then frozen while only the epinet parameters are optimized. In the harmonic-oscillator and chemical-motor examples, the epinets augment all thermodynamic blocks. In the 1D viscoplastic example, they augment only the non-kinetic contribution to the energy density and the entropy and dissipation potential densities,
while the canonical Poisson operator remains fixed. The same architecture is used for all reference distributions considered in each example. Each sampled epistemic index produces one rollout sample, and the resulting samples are used to compute the predictive mean, empirical standard deviation, and proper scoring rules.

\paragraph{Deep ensembles}
For deep ensembles, the hidden width cycles through the values in Table~\ref{tab:nn_hyperparameters} and remains fixed across the hidden layers of each member. Every member uses the full training dataset, with diversity introduced through the random seed and hidden width. The wall times in Tables~\ref{tab:harmonic_cost_and_scores},~\ref{tab:chemical_motor_cost_and_scores}, and~\ref{tab:viscoplastic_cost_and_scores} report the measured computational cost without parallelization. The predictive statistics and proper scoring rules are computed from the 50 trained members.

\paragraph{MC dropout}
MC dropout uses the same N-GENNs backbone as the deterministic model, with dropout active during training and inference. A dropout realization is held fixed over all timesteps of a rollout, so each sample follows one sampled vector field. We note that the 1D viscoplastic example uses a lower dropout rate of 0.1, rather than the rate of 0.2 used in the two ODE examples, because the higher rate produced non-convergent training runs and unstable rollout predictions.

\begin{table}[htbp]
    \caption{Architectural and training settings for the harmonic oscillator, idealized chemical motor, and 1D viscoplastic examples. Each method block includes the optimizer, learning-rate settings, network architecture, and method-specific training and inference settings. All neural networks are trained with Softplus activation functions.}
    \label{tab:nn_hyperparameters}
    \centering
    \scriptsize
    \setlength{\tabcolsep}{3pt}
    \renewcommand{\arraystretch}{1.08}
    \begin{tabular}{@{}>{\raggedright\arraybackslash}p{0.13\textwidth}
                    >{\raggedright\arraybackslash}p{0.30\textwidth}
                    >{\centering\arraybackslash}p{0.16\textwidth}
                    >{\centering\arraybackslash}p{0.16\textwidth}
                    >{\centering\arraybackslash}p{0.16\textwidth}@{}}
        \toprule
        \textbf{Method} & \textbf{Setting} & \textbf{Harmonic oscillator} & \textbf{Chemical motor} & \textbf{1D viscoplastic model} \\
        \midrule
        \multirow{12}{*}{\textit{S-PENNs}}
        & Optimizer & SOAP & SOAP & SOAP \\
        & Learning rate (base networks) & $3\times10^{-3}$ & $3\times10^{-3}$ & $3\times10^{-3}$ \\
        & Learning rate (epinets) & $3\times10^{-3}$ & $3\times10^{-3}$ & $3\times10^{-3}$ \\
        & Num. of hidden layers per block (base networks) & 2 & 2 & 2 \\
        & Num. of hidden layers per block (epinets) & 2 & 2 & 2 \\
        & Hidden-layer width (base networks) & 30 & 30 & 30 \\
        & Hidden-layer width (epinets) & 10 & 10 & 10 \\
        & Training epochs (base networks) & 10000 & 10000 & 10000 \\
        & Training epochs (epinets) & 200 & 2500 & 1000 \\
        & Epistemic-index dimension & 5 & 5 & 5 \\
        & Prior scale & 0.1 & 0.1 & 0.1 \\
        \addlinespace[2pt]
        \midrule
        \multirow{6}{*}{\textit{Deep ensembles}}
        & Optimizer & SOAP & SOAP & SOAP \\
        & Learning rate & $3\times10^{-3}$ & $3\times10^{-3}$ & $3\times10^{-3}$ \\
        & Ensemble members & 50 & 50 & 50 \\
        & Num. of hidden layers per block & 2 & 2 & 2 \\
        & Hidden-layer width choices & (20, 30, 40) & (20, 30, 40) & (20, 25, 30) \\
        & Training epochs per member & 10000 & 10000 & 10000 \\
        \addlinespace[2pt]
        \midrule
        \multirow{7}{*}{\textit{MC dropout}}
        & Optimizer & SOAP & SOAP & SOAP \\
        & Learning rate & $3\times10^{-3}$ & $3\times10^{-3}$ & $3\times10^{-3}$ \\
        & Dropout rate & 0.2 & 0.2 & 0.1 \\
        & Num. of hidden layers per block & 2 & 2 & 2 \\
        & Hidden-layer width & 30 & 30 & 30 \\
        & Training epochs & 10000 & 10000 & 10000 \\
        \bottomrule
    \end{tabular}
\end{table}

\section{Epinet construction for inverse problems}
\label{sec:epinet_inverse_problems}
The input-independent epinet branch introduced in Section~\ref{sec:epinet_input_independent} can also be used for inverse problems with unknown global physical parameters. These benchmarks are reported in the appendix because the main S-PENNs construction is developed for GENERIC dynamics, whereas the inverse problems isolate the same global-parameter mechanism in a PINN setting.

Let $\mathbf{s}$ denote the independent variables, with $\mathbf{s}=t$ for an ODE trajectory and $\mathbf{s}=(x,t)$ for a spatiotemporal PDE field, and let $\Omega$ be the domain on which the governing residual is evaluated. 
We consider inverse problems in which the state $u(\mathbf{s})$ and the unknown global physical parameters $\bm{\beta}\in\mathbb{R}^{d_\beta}$ are inferred from measurements while violations of the governing equations are penalized through the residual~\cite{raissi2019physics},
\begin{equation}
    \mathcal{R}(u;\bm{\beta})(\mathbf{s}) = 0,
    \qquad \mathbf{s}\in\Omega.
\end{equation}
Here, $\mathcal{R}$ denotes the governing residual. 
In the deterministic PINN baseline, the state is parameterized by a neural network 
$\mu_{\bm{\psi}}^{u}(\mathbf{s})$, and the global parameters are represented by the trainable vector $\bm{\beta}_{\bm{\psi}}$; both are included in the deterministic parameter set $\bm{\psi}$. The corresponding loss combines 
data-misfit and residual terms, with additional terms included when initial or boundary conditions are applicable.

Following the ENN notation in Eq.~\eqref{eq:enn_general}, the deterministic PINN is first pretrained and then held fixed while the epinet parameters $\bm{\phi}$ are optimized. The state-branch correction $\sigma_{\bm{\phi}}^{u}$ takes as input the stop-gradient feature vector $\bar{\bm{h}}_{\bm{\psi}}^{u}(\mathbf{s})=[\mathrm{sg}(\bm{h}_{\bm{\psi}}^{u}(\mathbf{s})),\mathbf{s}]$, which concatenates the frozen base features with the input coordinates. The parameter-branch correction $\bm{\sigma}_{\bm{\phi}}^{\beta}$ is deliberately input independent and depends only on the shared epistemic index $\mathbf{z}$. Each realization therefore assigns one global perturbation of the parameter vector over $\Omega$, rather than a coordinate-dependent parameter field. The augmented model is
\begin{equation}
    \label{eq:inverse_epinet_prediction}
    u_{\bm{\vartheta}}(\mathbf{s},\mathbf{z})
    =
    \mu_{\bm{\psi}}^{u}(\mathbf{s})
    +
    \sigma_{\bm{\phi}}^{u}(\bar{\bm{h}}_{\bm{\psi}}^{u}(\mathbf{s}),\mathbf{z}),
    \qquad
    \bm{\beta}_{\bm{\vartheta}}(\mathbf{z})
    =
    \bm{\beta}_{\bm{\psi}}
    +
    \bm{\sigma}_{\bm{\phi}}^{\beta}(\mathbf{z}),
\end{equation}
where $\bm{\vartheta}=(\bm{\psi},\bm{\phi})$. The state-branch correction follows the input-dependent epinet construction in Eqs.~\eqref{eq:epinet_decomposition}--\eqref{eq:epinet_prior_network} and the parameter-branch correction follows the same linear input-independent form as in Eq.~\eqref{eq:n-genn_global_matrix_head},
\begin{equation}
    \label{eq:inverse_linear_param_head}
    \begin{aligned}
        \bm{\sigma}_{\bm{\phi}}^{\beta}(\mathbf{z})
        &=
        \bm{\sigma}_{\bm{\phi}}^{\beta,\mathrm{learn}}(\mathbf{z})
        +
        w\,\bm{\sigma}^{\beta,\mathrm{prior}}(\mathbf{z}),\\
        \bm{\sigma}_{\bm{\phi}}^{\beta,\mathrm{learn}}(\mathbf{z})
        &=
        \sum_{n=1}^{d_z}z_n\,\bm{\phi}_n^{\beta}, \qquad
        \bm{\sigma}^{\beta,\mathrm{prior}}(\mathbf{z})
        =
        \sum_{n=1}^{d_z}z_n\,\bm{\zeta}_{n}^{\beta}.
    \end{aligned}
\end{equation}
Here, $\bm{\phi}_n^\beta\in\mathbb{R}^{d_\beta}$ are learnable coefficient vectors and $\bm{\zeta}_n^\beta\in\mathbb{R}^{d_\beta}$ are fixed prior coefficient vectors. For each sampled $\mathbf{z}$, the residual $\mathcal{R}$ is evaluated using the corresponding state realization $u_{\bm{\vartheta}}(\cdot,\mathbf{z})$ and global parameter vector $\bm{\beta}_{\bm{\vartheta}}(\mathbf{z})$. The shared epistemic index couples state and parameter uncertainty, while $\bm{\beta}_{\bm{\vartheta}}(\mathbf{z})$ remains a single global parameter vector over $\Omega$.

We evaluate this construction on two benchmark inverse problems from~\cite{zou2024neuraluq}, using the datasets distributed with the open-source NeuralUQ repository (\url{https://github.com/Crunch-UQ4MI/neuraluq}). The B-PINNs-HMC baseline~\cite{yang2021b} is reproduced from the same NeuralUQ implementation used in~\cite{zou2024neuraluq}. The proposed construction uses the two-branch epinet design described above, with the state-branch associated with $u$ and the parameter-branch associated with $\bm{\beta}$. Table~\ref{tab:inverse_epinet_settings} summarizes the architecture, training, and inference settings for the two benchmarks shown in the following sections. At inference, uncertainty is summarized by the empirical predictive mean and standard deviation with the same number of samples generated in the B-PINNs-HMC baselines for fair comparison. 

\begin{table}[htbp]
    \caption{Architecture, training, and inference settings for the proposed two-branch epinet construction in the Kraichnan--Orszag (KO) and Korteweg--de Vries (KdV) inverse problems. Tanh is used as the activation function and AdamW is the optimizer used for both training stages.}
    \label{tab:inverse_epinet_settings}
    \centering
    \small
    \setlength{\tabcolsep}{6pt}
    \renewcommand{\arraystretch}{1.08}
    \begin{tabular}{@{}>{\raggedright\arraybackslash}p{0.44\textwidth}
                    >{\centering\arraybackslash}p{0.23\textwidth}
                    >{\centering\arraybackslash}p{0.23\textwidth}@{}}
        \toprule
        \textbf{Setting} & \textbf{KO} & \textbf{KdV} \\
        \midrule
        Deterministic PINN hidden-layer widths & $(50,50)$ & $(50,50)$ \\
        Epinet hidden-layer widths (learnable and prior) & $(10,10)$ & $(10,10)$ \\
        Epistemic-index dimension & 5 & 5 \\
        Learning rate (base PINN / epinet) & $10^{-3}$ / $10^{-3}$ & $10^{-3}$ / $10^{-3}$ \\
        Training epochs (base PINN / epinet) & $20{,}000$ / $10{,}000$ & $20{,}000$ / $10{,}000$ \\
        Prior scale & 1.0 & 1.0 \\
        Reference distribution & $\mathcal{N}(\mathbf{0},\mathbf{I})$ & $\mathcal{N}(\mathbf{0},\mathbf{I})$ \\
        Predictive samples & 1000 & 500 \\
        \bottomrule
    \end{tabular}
\end{table}


\subsection{Inverse Kraichnan--Orszag system}
\label{sec:inverse_ko_system}
As a first benchmark, we consider the inverse Kraichnan--Orszag system, a three-variable nonlinear dynamical model arising from interactions among inviscid shear waves~\cite{wan2006multi}. The governing dynamics and initial conditions considered are
\begin{align}
    \dot{x}_1 &= a x_2 x_3, \label{eq:ko_ode}\\
    \dot{x}_2 &= b x_1 x_3, \nonumber\\
    \dot{x}_3 &= -(a+b)x_1x_2, \nonumber
\end{align}
\begin{equation}
    \label{eq:ko_ic}
    \bigl(x_1(0),x_2(0),x_3(0)\bigr)=(1.0,0.8,0.5).
\end{equation}
The reference coefficients are $a=1$ and $b=1$. In the inverse setting, both coefficients and the initial conditions are treated as unknowns. The task is to infer $a$ and $b$ together with the full trajectories $x_1$, $x_2$, and $x_3$ over $t \in [0,10]$ from sparse noisy measurements. 
Following~\cite{zou2024neuraluq}, we use 11 measurements 
for $x_1$ and $x_3$ and 7 measurements 
for $x_2$, each corrupted by zero-mean Gaussian noise with known standard deviation $0.05$. 

\begin{figure}[htbp]
    \centering
    \includegraphics[width=\textwidth]{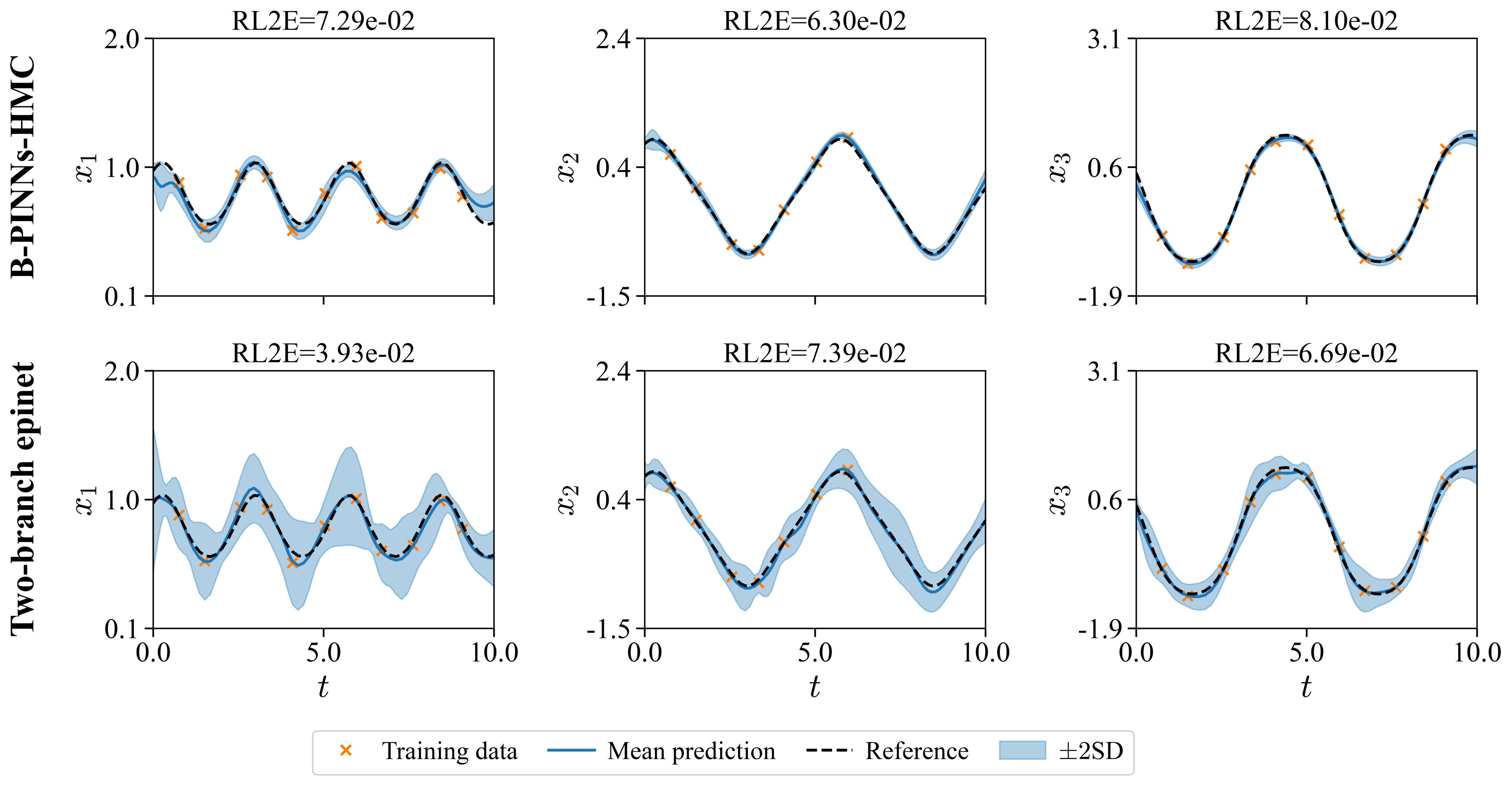}
    \caption{Predictive trajectories for the inverse Kraichnan--Orszag problem. Columns correspond to $x_1$, $x_2$, and $x_3$; rows compare B-PINNs-HMC and the proposed two-branch epinet construction. In each panel, the predictive mean is shown as a solid line, the reference solution as a dashed line, training measurements as orange markers, and the $\pm 2$ standard deviation band as a blue shaded region. Titles report the relative $\ell^2$ error (RL2E).}
    \label{fig:ko_solution_results}
\end{figure}

\begin{figure}[htbp]
    \centering
    \includegraphics[width=0.8\textwidth]{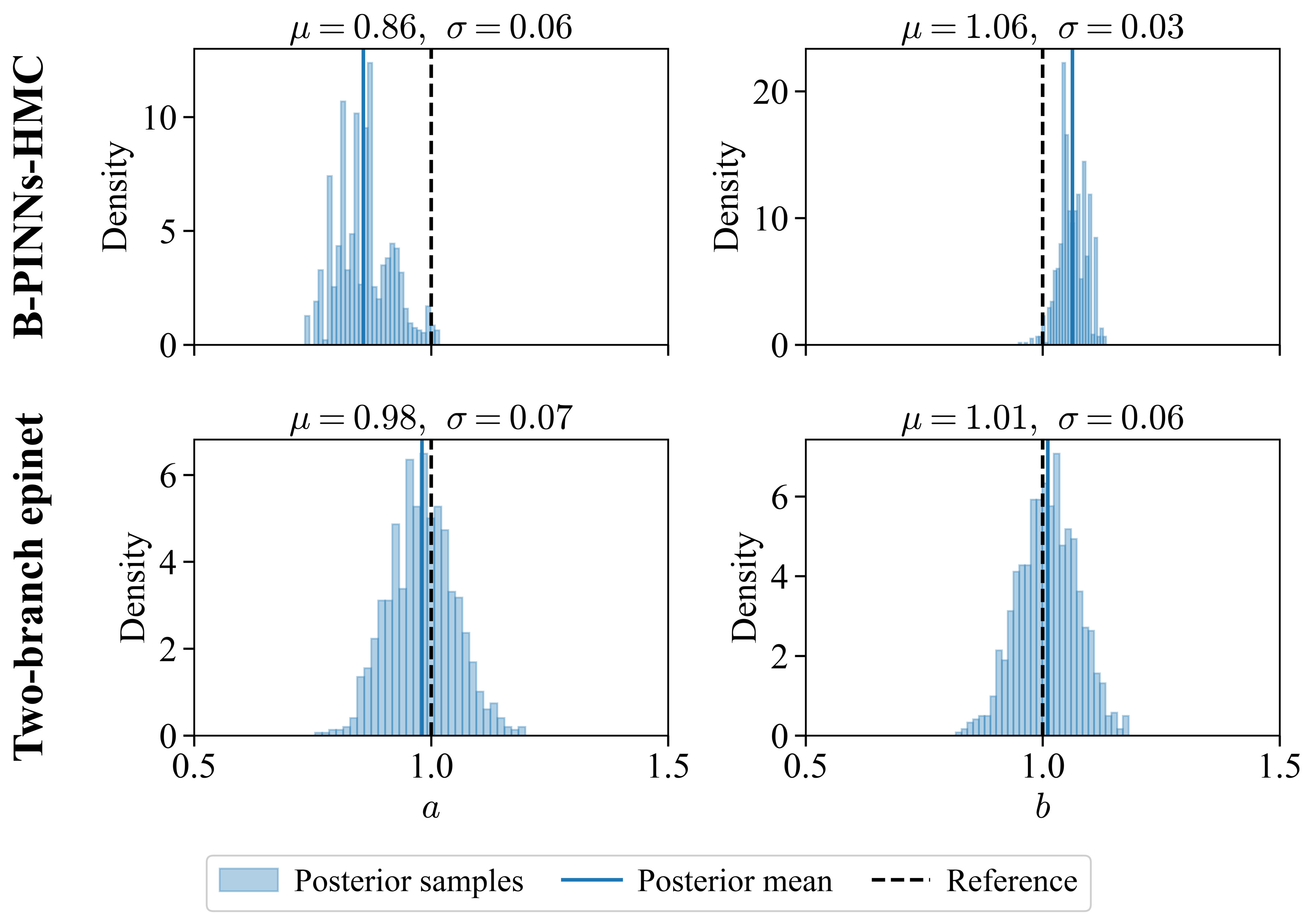}
    \caption{Estimated distributions of the unknown parameters for the inverse Kraichnan--Orszag problem. Columns correspond to $a$ and $b$; rows compare B-PINNs-HMC and the proposed two-branch epinet construction. Each panel shows the sample histogram, the sample mean as a solid vertical line, and the reference value as a dashed vertical line. The sample mean $\mu$ and standard deviation $\sigma$ are annotated above each panel.}
    \label{fig:ko_parameter_results}
\end{figure}

Fig.~\ref{fig:ko_solution_results} compares the reconstructed trajectories. Both the reproduced B-PINNs-HMC baseline and the proposed two-branch epinet construction recover the oscillatory dynamics from sparse noisy measurements, and their predictive means remain close to the reference trajectories over the full time interval. The componentwise relative $\ell^2$ errors show comparable state-reconstruction performance. The proposed construction reduces the errors for $x_1$ and $x_3$, from $7.29\times10^{-2}$ to $3.93\times10^{-2}$ and from $8.10\times10^{-2}$ to $6.69\times10^{-2}$, respectively, while B-PINNs-HMC is more accurate for $x_2$, with an error of $6.30\times10^{-2}$ compared to $7.39\times10^{-2}$. The estimated parameter distributions in Fig.~\ref{fig:ko_parameter_results} show that B-PINNs-HMC gives sample means and standard deviations of $a=0.86 \pm 0.06$ and $b=1.06 \pm 0.03$, whereas the proposed two-branch epinet construction gives more accurate results with $a=0.98 \pm 0.07$ and $b=1.01 \pm 0.06$.

\subsection{Inverse Korteweg--de Vries problem}
\label{sec:inverse_kdv}
The second benchmark considers an inverse Korteweg--de Vries (KdV) problem. Compared with the Kraichnan--Orszag system, this example tests the proposed construction on a field-valued inverse problem with localized wave profiles and soliton interaction. The KdV equation is a classical dispersive model for nonlinear wave propagation~\cite{benes2006decompositions}, and we write it as
\begin{equation}
    \label{eq:kdv_equation}
    u_t = \kappa_1 u u_x + \kappa_2 u_{xxx} + f,
\end{equation}
where the reference problem uses $\kappa_1 = 1.5$, $\kappa_2 = 0.25$ and $f=0$. 
Following~\cite{benes2006decompositions}, the exact two-soliton solution is
\begin{equation}
    \label{eq:kdv_exact_solution}
    u(x,t)=2\partial_x^2\log\left[
    \exp(-\omega_1-\omega_2) + \exp(\omega_1-\omega_2)
    + \exp(\omega_2-\omega_1)
    + \frac{(a_1-a_2)^2}{(a_1+a_2)^2}\exp(\omega_1+\omega_2)
    \right],
\end{equation}
where the phase variables are
\begin{align}
    \omega_i = a_i x + a_i^3 \chi t + b_i, \qquad i=1,2,
\end{align}
with $a_1=1$, $a_2=2$, $b_1=b_2=\log(3)/2$, and $\chi=1$. 

\begin{figure}[htbp]
    \centering
    \includegraphics[width=0.8\textwidth]{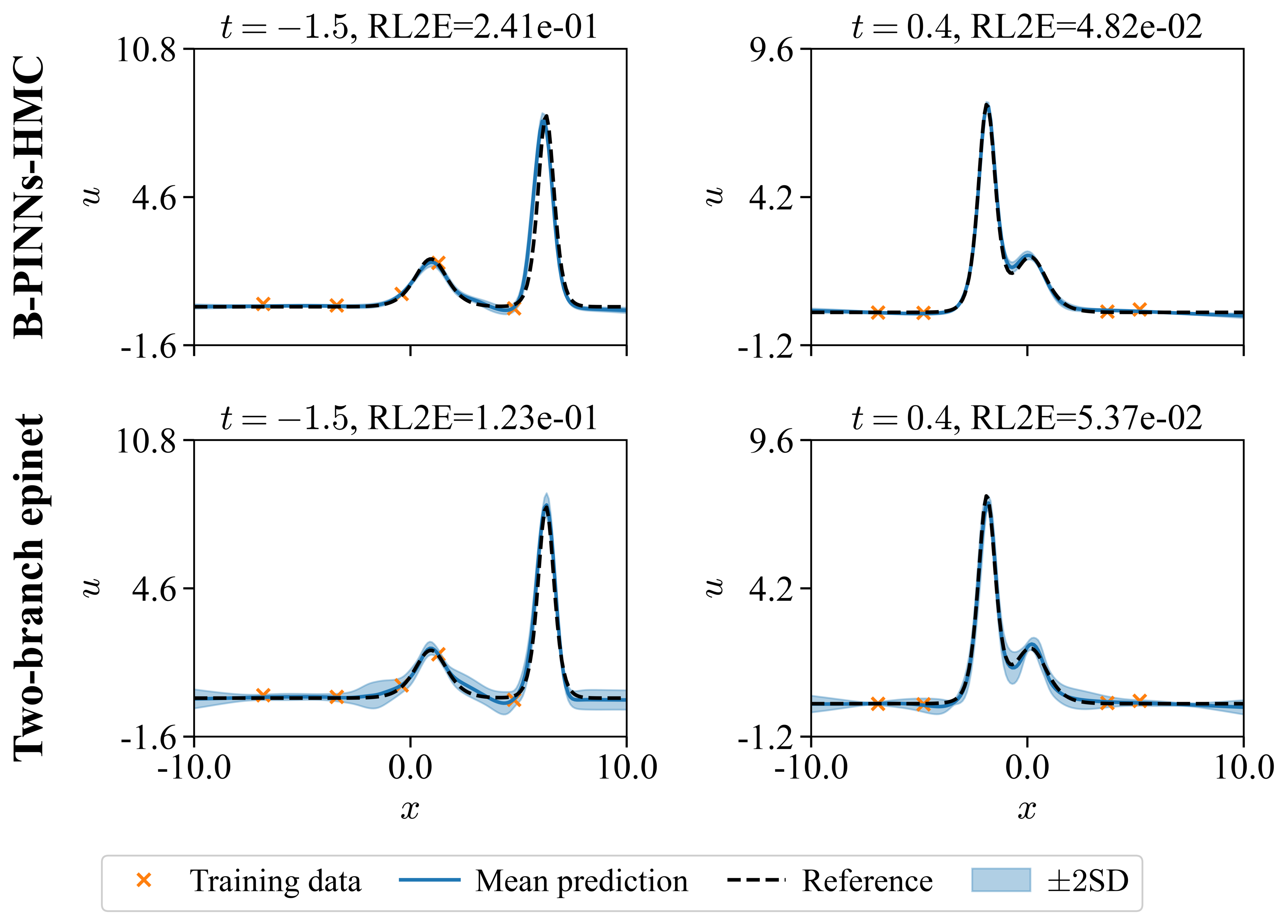}
    \caption{Predictive solution profiles for the inverse KdV problem at two representative time slices. Columns correspond to $t=-1.5$ and $t=0.4$; rows compare B-PINNs-HMC and the proposed two-branch epinet construction. In each panel, the predictive mean is shown as a solid line, the reference solution as a dashed line, training measurements as orange markers, and the $\pm 2$ standard deviation band as a blue shaded region. Titles report the relative $\ell^2$ error (RL2E). The scaling for $t$ follows~\cite{benes2006decompositions}.}
    \label{fig:kdv_solution_results}
\end{figure}

\begin{figure}[htbp]
    \centering
    \includegraphics[width=0.8\textwidth]{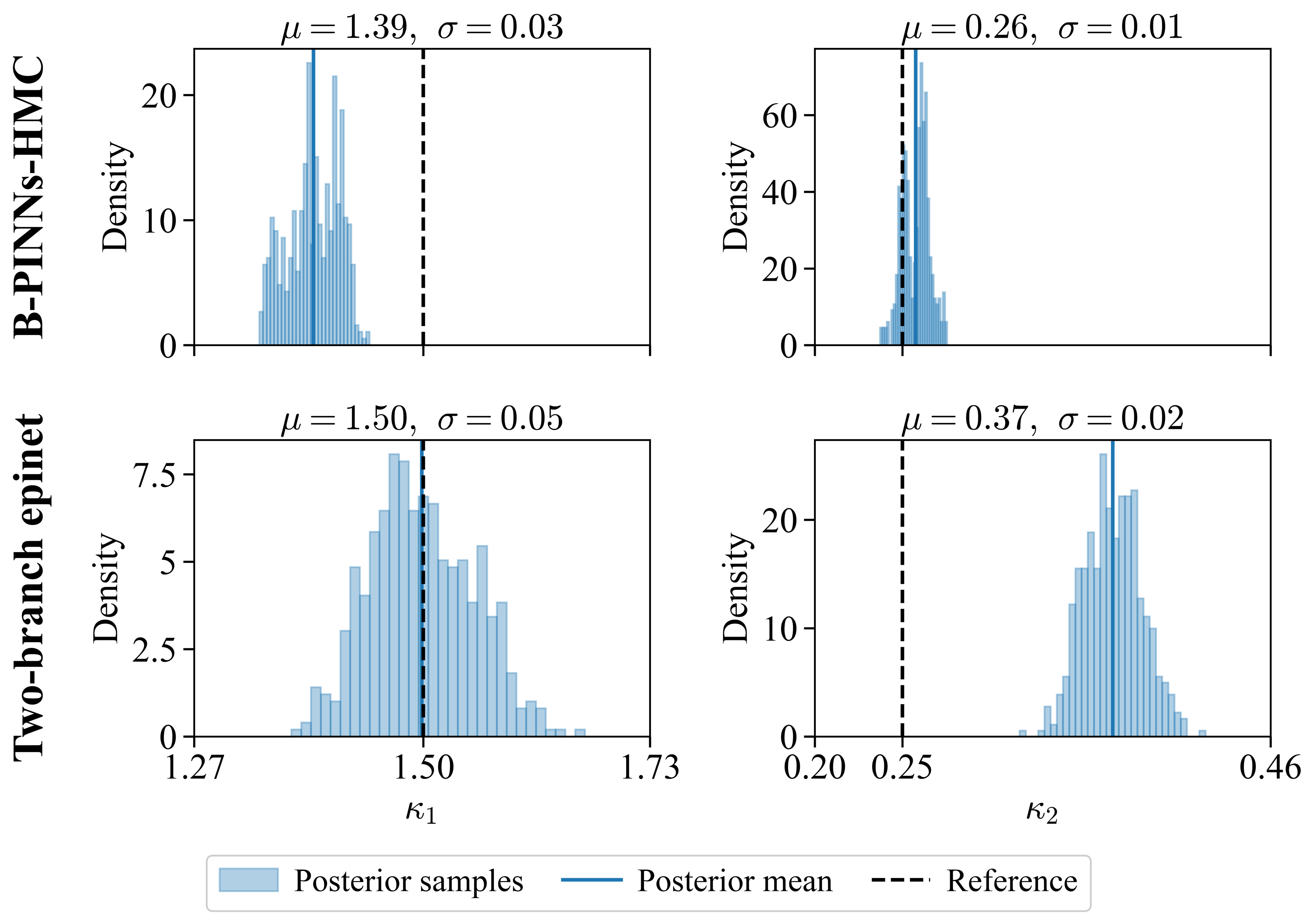}
    \caption{Estimated distributions of the unknown parameters for the inverse KdV problem. Columns correspond to $\kappa_1$ and $\kappa_2$; rows compare B-PINNs-HMC and the proposed two-branch epinet construction. Each panel shows the sample histogram, the sample mean as a solid vertical line, and the reference value as a dashed vertical line. The sample mean $\mu$ and standard deviation $\sigma$ are annotated above each panel.}
    \label{fig:kdv_parameter_results}
\end{figure}

The inverse task is to infer $\kappa_1$ and $\kappa_2$ while reconstructing $u(x,t)$ over the spatiotemporal domain from sparse noisy measurements. We again use the data from~\cite{zou2024neuraluq}, with $200$ random measurements of $u$ and $100$ measurements of $f$. Both measurement types are perturbed by zero-mean Gaussian noise with known standard deviation $0.1$. Fig.~\ref{fig:kdv_solution_results} reports the predicted profiles at two representative time slices. Both B-PINNs-HMC and the proposed two-branch epinet construction recover the dominant two-soliton structure, including the narrow high-amplitude peak and the smaller secondary wave. Their errors differ by regime. At $t=-1.5$, where the soliton peaks remain well separated, the proposed construction reduces the relative $\ell^2$ error from $2.41\times10^{-1}$ to $1.23\times10^{-1}$. At $t=0.4$, closer to the interaction regime, B-PINNs-HMC gives the smaller error, $4.82\times10^{-2}$ compared with $5.37\times10^{-2}$ for the proposed two-branch epinet construction. The parameter distributions in Fig.~\ref{fig:kdv_parameter_results} show that B-PINNs-HMC gives $\kappa_1 = 1.39 \pm 0.03$ and $\kappa_2 = 0.26 \pm 0.01$, whereas the proposed two-branch epinet construction gives $\kappa_1 = 1.50 \pm 0.05$ and $\kappa_2 = 0.37 \pm 0.02$. Thus, the input-independent parameter branch identifies $\kappa_1$ nearly exactly, but it overestimates $\kappa_2$. In contrast, B-PINNs-HMC is closer for $\kappa_2$ but biased low for $\kappa_1$.

\section*{Code availability}
The code will be made available upon publication.

\section*{Declaration of competing interest}
The authors declare that they have no known competing financial interests or personal relationships that could have appeared to influence the work reported in this paper.

\section*{Acknowledgments}
The authors acknowledge support from the US Department of the Army W911NF2310230.

\section*{Declaration of generative AI and AI-assisted technologies in the writing process}
During the preparation of this work, the authors used ChatGPT to check grammar and improve sentence clarity. After using this tool, the authors reviewed and edited the content as needed and took full responsibility for the final published article.

\bibliographystyle{unsrt}
\bibliography{references}

\end{document}